\documentclass[journal]{IEEEtran}
\usepackage{amsmath,amssymb,amsfonts}

\usepackage[colorlinks,linkcolor=blue,anchorcolor=blue,citecolor=blue]{hyperref}
\usepackage{array}
\usepackage[caption=false,font=normalsize,labelfont=sf,textfont=sf]{subfig}
\usepackage{textcomp}
\usepackage{stfloats}
\usepackage{url}
\usepackage{verbatim}
\usepackage{graphicx}
\usepackage[ruled, linesnumbered]{algorithm2e}
\usepackage{multirow}
\usepackage{booktabs}
\usepackage{diagbox}
\usepackage{wrapfig}
\usepackage{slashbox}
\usepackage{arydshln}

\def\BibTeX{{\rm B\kern-.05em{\sc i\kern-.025em b}\kern-.08em
    T\kern-.1667em\lower.7ex\hbox{E}\kern-.125emX}}
\usepackage{balance}

\begin{document}
\title{Long-Tailed Classification Based on Coarse-Grained Leading Forest and Multi-center Loss}
\author{
        Jinye~Yang,
        Ji~Xu$^{*}$,~\IEEEmembership{Member,~IEEE},
       Di~Wu,~\IEEEmembership{Member,~IEEE},
       Jianhang~Tang,~\IEEEmembership{Member,~IEEE},
       Shaobo~Li,
        Guoyin~Wang$^{*}$,~\IEEEmembership{Senior Member,~IEEE}
%
%
%
\IEEEcompsocitemizethanks{
This work has been supported by the National Key Research and Development Program of China under grant 2020YFB1713300, the National Natural Science Foundation of China under grants 62366008, 61966005, and 62221005, Guizhou Provincial Basic Research Program (Natural Science) (No. ZK[2024]YiBan048), Youth Science And Technology Talent Growth Project of Guizhou Education Department (Grant No. QianjiaoJi[2024]22).
\IEEEcompsocthanksitem \emph{*: Corresponding author.}
\IEEEcompsocthanksitem Jinye Yang, Ji Xu, Shaobo Li and Jianhang Tang are with State Key Laboratory of Public Big Data, Guizhou University, Guiyang 550025, China.
\protect E-mail: {jixu@gzu.edu.cn; gs.jinyeyang21@gzu.edu.cn; jhtang@gzu.edu.cn; lishaobo@gzu.edu.cn}
\IEEEcompsocthanksitem Di Wu is the College of Computer and Information Science, Southwest University, Chongqing, China 400715, China.
\protect E-mail: {wudi1986@swu.edu.cn}
\IEEEcompsocthanksitem Guoyin Wang is with Chongqing Key Laboratory of Computational Intelligence, Chongqing University of Posts and Telecommunications, Chongqing 400065, China
\protect E-mail: {wanggy@ieee.org}
}
\thanks{Manuscript received XXXX, 2023.}}

\markboth{IEEE Transactions on Emerging Topics in Computational Intelligence, July ~2024 (Accepted Manuscript)}%
{Long-Tailed Classification Based on Coarse-Grained Leading Forest and Multi-center Loss }

\IEEEtitleabstractindextext{
\begin{abstract}
Long-tailed (LT) classification is an unavoidable and challenging problem in the real world. Most existing long-tailed classification methods focus only on solving the class-wise imbalance while ignoring the attribute-wise imbalance. The deviation of a classification model is caused by both class-wise and attribute-wise imbalance. Due to the fact that attributes are implicit in most datasets and the combination of attributes is complex, attribute-wise imbalance is more difficult to handle. For this purpose, we proposed a novel long-tailed classification framework, aiming to build a multi-granularity classification model by means of invariant feature learning. This method first unsupervisedly constructs Coarse-Grained Leading Forest (CLF) to better characterize the distribution of attributes within a class. Depending on the distribution of attributes, one can customize suitable sampling strategies to construct different imbalanced datasets. We then introduce multi-center loss (MCL) that aims to gradually eliminate confusing attributes during feature learning process. The proposed framework does not necessarily couple to a specific LT classification model structure and can be integrated with any existing LT method as an independent component. Extensive experiments show that our approach achieves state-of-the-art performance on both existing benchmarks ImageNet-GLT and MSCOCO-GLT and can improve the performance of existing LT methods. Our codes are available on GitHub: \url{https://github.com/jinyery/cognisance}\footnote{\color{red}\copyright 2024 IEEE.  Personal use of this material is permitted.  Permission from IEEE must be obtained for all other uses, in any current or future media, including reprinting/republishing this material for advertising or promotional purposes, creating new collective works, for resale or redistribution to servers or lists, or reuse of any copyrighted component of this work in other works.}
\end{abstract}

}

\maketitle

\IEEEdisplaynontitleabstractindextext

\IEEEpeerreviewmaketitle

\section{Introduction}
\IEEEPARstart{I}{n} real-world applications, training samples typically exhibit a long-tailed distribution, especially for large-scale datasets \cite{zhang2023deep, yang2022multi, chaudhary2022real}. Long-tailed distribution means that a small number of head categories contain a large number of samples, while the majority of tail categories have a relatively limited number of samples. This imbalance can lead to traditional classification algorithms preferring to handle head categories and performing poorly when dealing with tail categories. Solving the problem of long-tailed classification is crucial as tail categories may carry important information such as rare diseases, critical events, or characteristics of minority groups \cite{buda2018systematic, ando2017deep, yang2019me, cao2019learning}. In order to effectively address this challenge, researchers have proposed various methods. However, what this article wants to emphasize is that the long-tailed distribution problem that truly troubles the industry is not all the class-wise long-tailed problem (that is currently the most studied). What truly hinders the success of machine learning methods in the industry is the attribute-wise long-tailed problem, such as the long-tailed distribution in unmanned vehicle training data, the weather distribution, day and night distribution \cite{sun2019see}, etc., which are not the targets predicted by the model but can affect the predicting performance if not properly addressed. Therefore, the long-tailed here is not only among classes, but also among attributes, where the attribute represents all the factors that cause attribute-wise changes, including object-level attributes (such as the specific vehicle model, brand, color, etc.) and image-level attributes (such as lighting, climate conditions, etc.), this emerging task is named {Generalized Long-Tailed Classification} (GLT) \cite{tang2022invariant}.

\begin{figure}[htb]
\centering
\scriptsize
\includegraphics[width=1.0\linewidth]{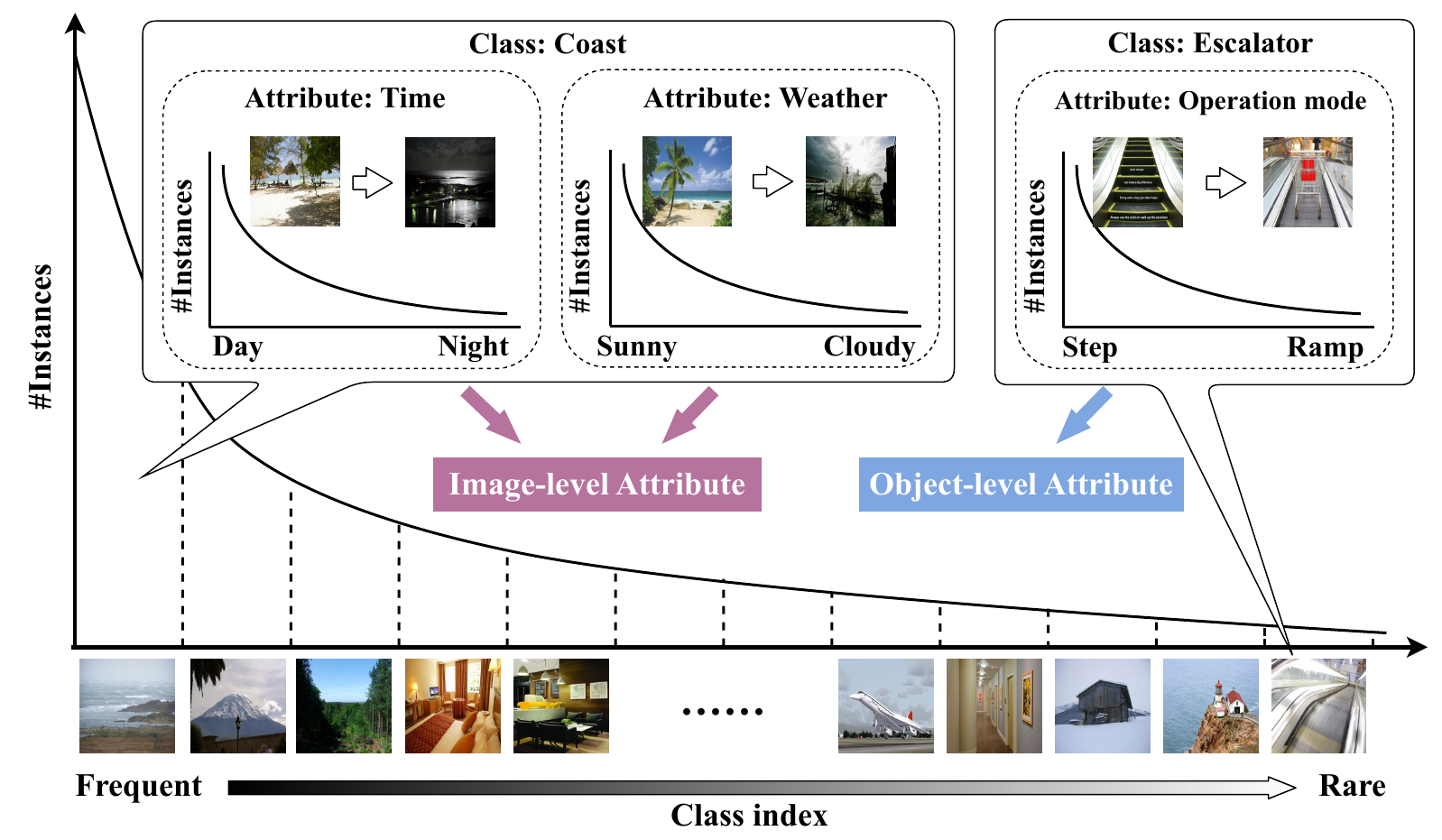}
\caption{Inter-class long-tailed distribution and attribute-wise long-tailed distribution.}
\label{fig:long-tail}
\end{figure}


In Fig. \ref{fig:long-tail}, it is evident that there is a class-wise imbalance among different categories, especially for the head class such as ``Coast", which has a much larger number of samples than the tail class such as ``Escalator". Apart from class-wise imbalance, there is a significant difference in the number of samples corresponding to different attributes within a category (termed as  attribute-wise imbalance for short). For example, in the category ``Coast", the number of samples during the day is much greater than that at night, and the number of samples on sunny days is also much greater than that on cloudy days. Even in the tail category ``Escalator", the number of samples in ``Step Escalator" is greater than that in ``Ramp Escalator".

\begin{figure}[htbp]
\centering
\scriptsize
\includegraphics[width=1.0\linewidth]{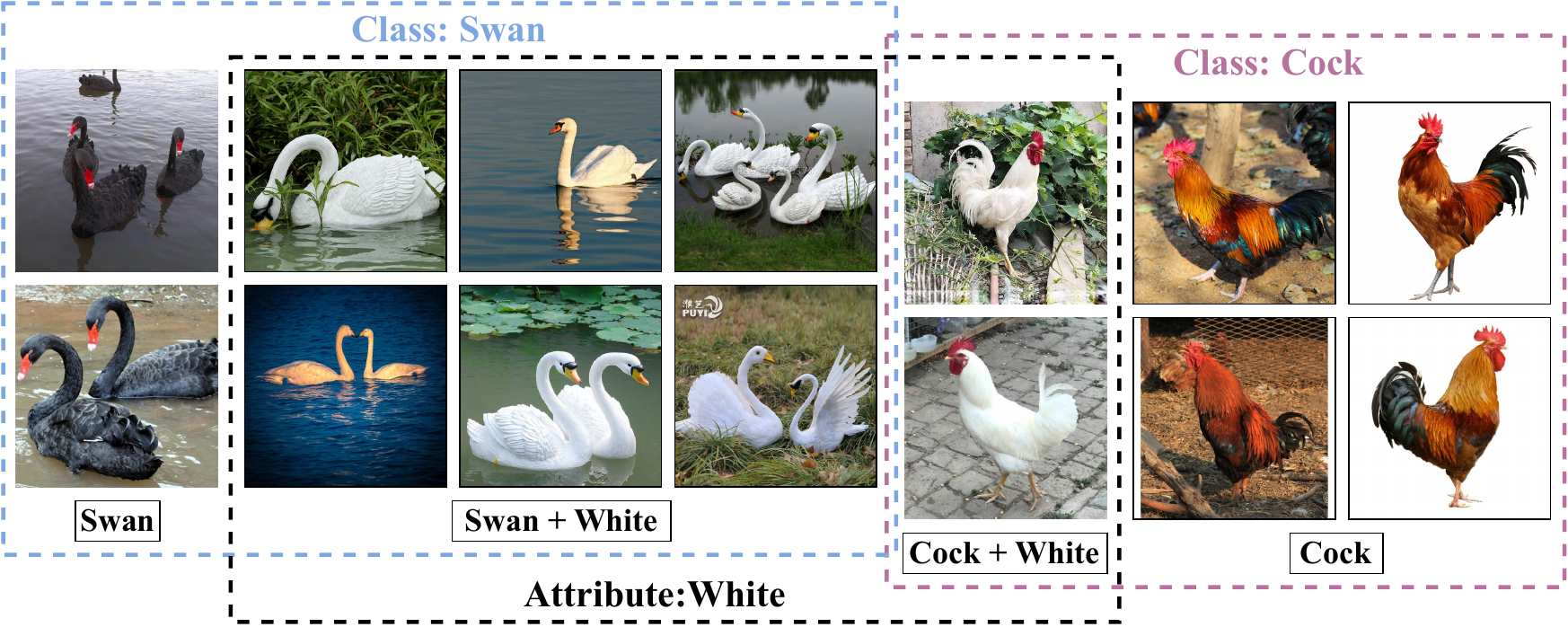}
\caption{Spurious correlation of the ``White" attribute with the ``Swan" category. Firstly, black swans are more likely to be misclassified than white swans in the category ``Swan", even though they both belong to the same category. Secondly, the attribute ``White" may be falsely correlated with ``Swan", so when ``White" appears in images of ``Cock", there is a high risk that the ``cock" will be misclassified as a ``Swan".}
\label{fig:inv_attr}
\end{figure}

Attribute-wise imbalance is fundamentally more difficult to handle than class-wise imbalance,
as shown in Fig. \ref{fig:inv_attr}.

Current research mostly focuses on solving the class-wise long-tailed problem, where resampling \cite{drummond2003c4} \cite{ han2005borderline}, loss reweighting \cite{he2009learning} \cite{ lin2017focal}, tail class augmentation \cite{li2024feature}, and Gaussian perturbed feature \cite{li2022long} \cite{li2024adjusting} are used to rebalance the training process. However, most of these methods sacrifice the performance of the head class to improve the performance of the tail class, which is like playing a performance seesaw, making it difficult to fundamentally improve the performance for all classes. Apart from resampling and reweighting, there are some methods (e.g., \cite{kang2019decoupling, zhou2020bbn}) believe that data imbalance does not affect feature learning, so the training of the model is divided into two stages of feature learning and classifier learning. However, this adjustment is only a trade-off between accuracy and precision \cite{tang2022invariant, zhu2022cross}, and the confusion areas of similar attributes in the feature space learned by the model do not change. As depicted in Fig \ref{fig:inv_attr}, it is the attribute-wise imbalance that leads to spurious correlations between the head attributes and a particular category. This means that these head attributes correspond to the spurious features (i.e., a confusing region in the feature space) of the category.

 To alleviate the performance-degrading affects from spurious features, Arjovsky \emph{et. al} proposed the concept of IRM \cite{arjovsky2020invariant}, for which the construction of different environments is a prerequisite for training. The challenge in this paper is to construct controllable environments with different attribute distributions. Environments with different category distributions are easy to construct because the labels are explicit, while the attributes are implicit for most datasets. Therefore, even if the class-wise imbalance is completely eliminated, attribute-wise imbalance still exists. At the same time, because of the nature that attributes can be continuously superimposed and combined, the boundaries of attributes are also complex, thus we design a sampling method based on unsupervised learning. The result of this unsupervised learning portrays the distribution of attributes within the same category and can control the granularity of the portraying of attributes according to the setting of hyperparameters.

\begin{figure}[htbp]
\centering
\includegraphics[width=0.5\linewidth]{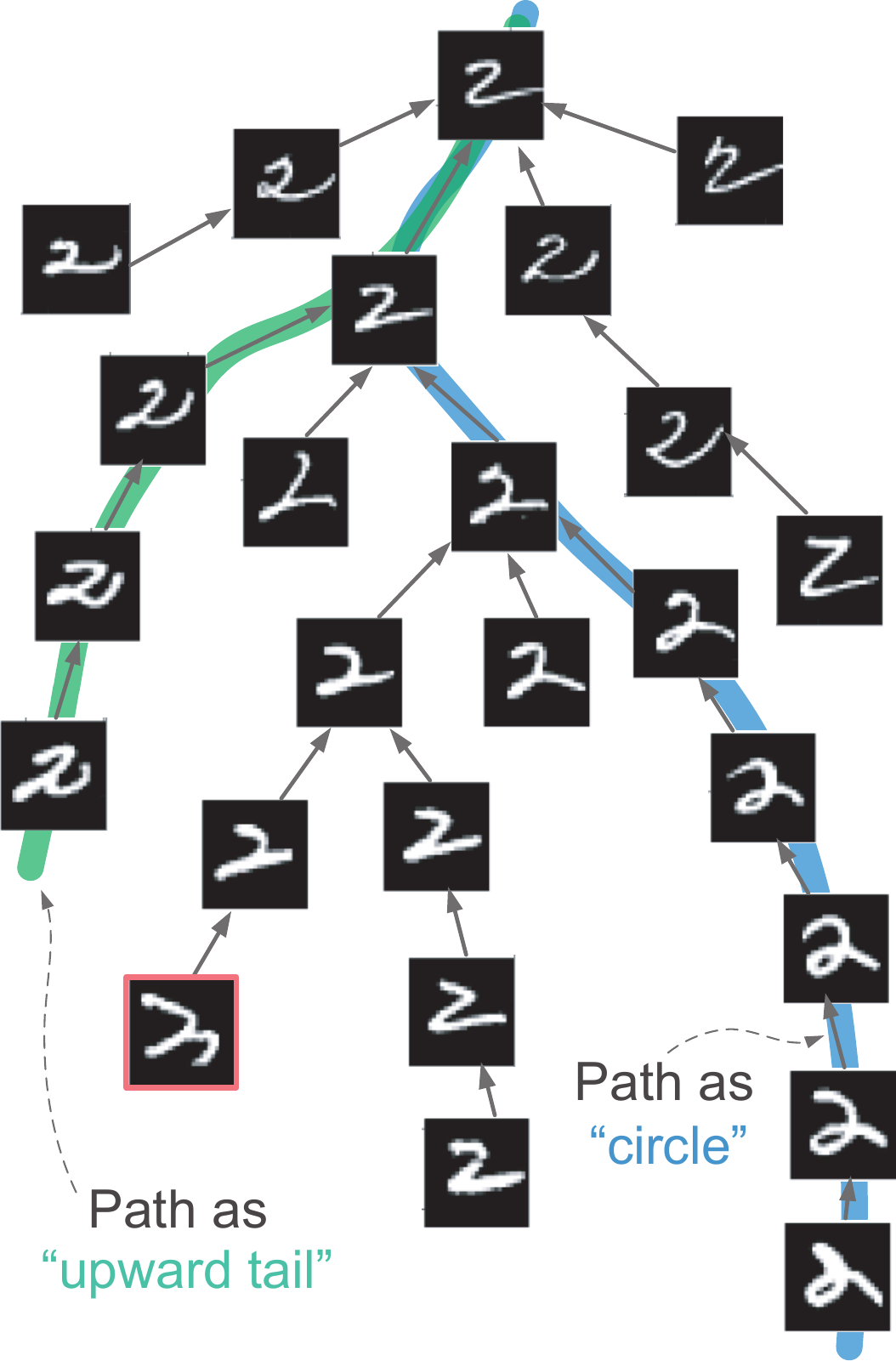}
\caption{A leading tree constructed from digit `2' in MNIST dataset (taken from our previous work \cite{xu2021lapoleaf}), in which each path can reflect an implicit attribute within the same class. Note that this figure is used only to explain our intuition and motivation, we did not evaluate our method on MNIST dataset.}
\label{fig:evolution1}
\end{figure}

Our motivation is based on a reasonable assumption that the differences between samples within the same category are the result of the gradual separation and evolution of attributes. This is observed in a leading tree ( Fig. \ref{fig:evolution1}) in our previous work \cite{xu2021lapoleaf}, where the gap from the root node to the leaf nodes doesn't happen all at once, but is the result of continuous evolution and branching. The evolution is not only reflected at a coarse-grained level, but also within the same category where the transitions are more subtle.
Within the same category of `2', the samples along each path evolve gradually. We can recognize the implicit attributes of ``upward tail" and ``circle" with human perception, although it is not explicitly described by the algorithm. Without gaining knowledge on the attribute-wise distribution with the same class, the existing GLT method uses the try-and-error methodology to find poorly predicted environments and train invariant features across them \cite{tang2022invariant}. In contrast, our method first find the samples that follow implicit attribute-wise evolution via unsupervised learning within each given class and construct effective environments.

%

  This paper proposes a framework termed as {\textsc{Cognisance}}, which is founded on \underline{Co}arse-\underline{g}rai\underline{n}ed Lead\underline{i}ng Fore\underline{s}t \underline{an}d {Multi-\underline{ce}nter Loss}. {\textsc{Cognisance}} handles class-wise and attribute-wise imbalance problems simultaneously by constructing different \emph{environments} \footnote{ Environments are the data subsets sampled from existing image datasets to reflect both class-wise and attribute-wise imbalance. }, in which we design a new sampling method based on a structure of samples called  {Coarse-Grained Leading Forest (CLF)}.  A CLF is constructed by unsupervised learning that characterizes the attribute distributions within a class and guides the data sampling in different environments during training process. In the experimental setup of this paper, two environments are constructed, one of which is the original environment without special treatment, and the distributions of categories and attributes in the other are balanced. In order to gradually eliminate confusing pseudo-features during the training process, we design a new metric learning loss, {Multi-Center Loss (MCL)}, which is inspired by \cite{tang2022invariant} and \cite{wen2016discriminative}, extends the center loss to its Invariant Risk Minimization (IRM) version. MCL enables our model to better learn invariant features and further improves the robustness when compared to its counterparts in \cite{tang2022invariant} and \cite{wen2016discriminative}. In addition, the proposed framework is not coupled with a specific backbone model or loss function, and can be seamlessly integrated into other LT methods for performance enhancement.

Our contributions can be summarized as follows:
\begin{itemize}
\item We designed a novel unsupervised learning-based sampling scheme based on CLF to guide the sampling of different environments in the IRM process. This scheme deals with class-wise and attribute-wise imbalance simultaneously.
\item We combined the idea of invariant feature learning and the property of CLF structure to design a new metric learning loss (MCL), which can enable the model to gradually remove the influence of pseudo-features during the training process. MCL improves the robustness of the model and takes into account both precision and accuracy of the prediction.
\item We conducted extensive experiments on two existing benchmarks, ImageNet-GLT and MSCOCO-GLT, to demonstrate the effectiveness of our framework on further improving the performance of popular LT lineups against \cite{tang2022invariant}.
\item {\color{black}We propose a novel framework to deal with noise in long-tailed datasets, in which we design a new loss function and a noise selection strategy based on CLF. The framework is validated on two long-tailed noise datasets, and achieves encouraging performance.}
\end{itemize}

The remainder of the paper is structured as follows. Section \ref{sec:RelatedWorks} briefly introduces the closely related preliminaries. Section \ref{sec:CognisanceFramework} gives detailed description of \textsc{Cognisance} framework. The experimental study is presented in Section \ref{sec:Experiment} and some discussion on the conceptualization and implementation of \textsc{Cognisance} are given in Section  \ref{sec:Discussion}. Finally, we reach the conclusion in Section \ref{sec:Conclusion}.

\section{Related Work}\label{sec:RelatedWorks}

\subsection{Long-Tailed Classification}

The key challenge of long-tailed classification is to effectively deal with the imbalance of data distribution to ensure that excellent classification performance can be achieved both between the head and the tail. Current treatments for long-tailed classification can be broadly categorized into three groups \cite{zhang2023deep}: 1) \underline{Class Re-balancing}, which is the mainstream paradigm in long-tailed learning, aims to enhance the influence of tail samples on the model by means of re-sampling \cite{wang2020devil, han2005borderline, estabrooks2004multiple, zhang2021learning}, re-weighting \cite{ren2020balanced, elkan2001foundations, jamal2020rethinking, tan2020equalization, tan2021equalization} or logit adjustment \cite{menon2020long, li2022long} during the model training process, and some of these methods \cite{kang2019decoupling, zhou2020bbn} consider that the unbalanced samples do not affect the learning of the features, and thus divide feature learning and classifier learning into two phases, and perform operations such as resampling only in the classifier learning phase. 2) \underline{Information Augmentation} based approaches seek to introduce additional information in model training in order to improve model performance in long-tailed learning. There are two approaches in this method type: migration learning \cite{chu2020feature, wang2021rsg, hu2020learning} and data augmentation \cite{cubuk2020randaugment, liu2020deep, zhang2017mixup}. 3) \underline{Module Improvement} seeks to improve network modules in long-tailed learning such as RIDE \cite{wang2020long} and TADE \cite{zhang2022self}, both of which deal with long-tailed recognition independent of test distribution by introducing integrated learning of multi-expert models in the network.{\color{black} In addition, SHIKE \cite{jin2023long} propose depth-wise knowledge fusion to fuse features between different shallow parts and the deep part in one network for each expert, which makes experts more diverse on representation.}

In addition, a recent study proposed the concept of Generalized Long-Tailed Classification (GLT) \cite{tang2022invariant}, which first recognized the problem of long-tailed of attributes within a class and pointed out that the traditional long-tailed classification methods represent the classification model as $p(Y|X)$. This can be further decomposed as $p(Y|X) \propto p(X|Y) \cdot p(Y)$, the formula identifies the cause of class bias as $p(Y)$. However, the distribution of $p(X|Y)$ also changes in different domains, so the classification model is extended to the form of Eq. (\ref{eq:glt}) based on the Bayesian Theorem in this study.
\begin{equation}
\label{eq:glt}
p(Y=k|z_c, z_a) = \frac{p(z_c|Y=k)}{p(z_c)}\cdot\underbrace{\frac{p(z_a|Y=k, z_c)}{p(z_a|z_c)}}_{attribute\ bias}\cdot \underbrace{p(Y=k)}_{class\ bias},
\end{equation}
where $z_c$ is the invariant present in the category, and the attribute-related variable $z_a$ is the domain-specific knowledge in different distributions. Taking the mentioned ``swan" as an example, the attribute ``color" of ``Swan" belongs to $z_a$, while the attributes of ``Swan" such as feathers and shape belong to $z_c$.

\textbf{Remark 1}: \emph{In practical applications, the formula does not impose the untangling assumption, i.e., it does not assume that a perfect feature vector $z=[z_c;z_a]$ can be obtained, where $z_a$ and $z_c$ are separated.}



\subsection{Invariant Risk Minimization}

The main goal of IRM \cite{arjovsky2020invariant} is to build robust learning models that can have the same performance on different data distributions. In machine learning, we usually hope that the trained model can perform well on unseen data, which is called risk minimization. However, in practice, there may be distributional differences between the training data and the test data, known as domain shift, which causes the model to perform poorly on new domains. The core idea of IRM is to solve the domain adaptation problem by encouraging models to learn features that are invariant across data domains. This means that the model should focus on those shared features that are present in all data domains rather than overfitting a particular data distribution. The objective of IRM is shown in (\ref{eq:irm}).

\begin{equation}
\label{eq:irm}
\begin{split}
&\mathop{\min}\limits_{\substack{\Phi:\mathcal{X}\rightarrow\mathcal{H} \\ w:\mathcal{H}\rightarrow\mathcal{Y}}}\sum\limits_{e\in \mathcal{E}_{tr}}R^e(w\circ\Phi)\\s.t. \quad &w\in\mathop{\arg\min}\limits_{\overline{w}:\mathcal{H}\rightarrow\mathcal{Y}}R^e(\overline{w}\circ\Phi),\ for\ all\ e\in\mathcal{E}_{tr},
\end{split}
\end{equation}
 where $\mathcal{E}_{tr}$ represents all training environments, $\mathcal{X}$, $\mathcal{H}$, and $\mathcal{Y}$ represent inputs, feature representations, and prediction results, respectively. $\Phi$ and $w$ are feature learner and classifier, respectively, and $R^e$ denotes the risk under the environment $e$, the goal of IRM is to find a general feature learning solution that can perform stably on all environments, thus improving the model's generalization ability.

\subsection{Optimal Leading Forest}

The CLF proposed in this paper starts with OLeaF \cite{xu2018LoDOG}, so we provide a brief introduction to its ideas and algorithms here. The concept of optimal leading forest originates from a clustering method based on density peak \cite{rodriguez2014clustering}, and the two most critical factors in the construction of OLeaF are the density of a data point and the distance of the data point to its nearest neighbor with higher densitiy. Let ${\boldsymbol I}=\left\{1,2,...,N\right\}$ be the index set of dataset $\mathcal{X}$, and $d_{i,j}$ represent the distance between data points $\boldsymbol{x}_i$ and $\boldsymbol{x}_j$ (any distance metric can be used), and let $\rho_i=\sum_{j\in \boldsymbol I \backslash\{i\}}\exp\big(-(d_{i,j}/d_c)^2\big)$ be the density of data point $\boldsymbol{x}_i$, where $d_c$ is the bandwidth parameter. If there exists $\xi_i=\arg\min_j\left\{ d_{i,j}|\rho_j>\rho_i \right\}$, then a tree structure (termed as \emph{leading tree}) can be established based on $\boldsymbol{\xi}=\{ \xi_i\}_{i=1}^{N}$. Let $\delta_i = d_{\xi_i, i}$, $\gamma_i=\rho_i\times\delta_i$, then the larger $\gamma_i$ represents the higher potential of the data point $\boldsymbol{x}_i$ to be selected as a cluster center. Intuitively, if an object $\boldsymbol{x}_i$ a has a large $\rho_i$, it means it has a lot of close neighbors; and if $\delta_i$ is large, it is far away from another data point with a larger $\rho$ value, so $\boldsymbol{x}_i$ has a good chance to become the center of a cluster. The root nodes (centers of clustering) can be selected based on the ordering of $\{\gamma_i\}$. In this way, an entire leading tree will be partitioned using the centers indicated by largest elements in $\{\gamma_i\}$, and the resultant collection of leading trees is called an Optimal Leading Forest (OLeaF).

The OLeaF structure has been observed has the capability of revealing the attribute-wise difference evolution \cite{xu2021lapoleaf}, so it can be employed to construct the environments in GLT training.  Although the unsupervised learning procedure dose not automatically tag the attribute label of each path, but we could recognize via human cognitive ability the meaningfulness of each path (e.g., the ``circle" attribute and ``upward tail" attribute in Fig. ).

\section{\textsc{Cognisance} framework}\label{sec:CognisanceFramework}
%

 In order to improve the generalization ability of the model on data with different category distributions and different attribute distributions, this paper proposes an framework named \textsc{Cognisance} based on invariant feature learning, which firstly uses coarse-grained leading forest to construct different environments considering both class-wise and attribute-wise imbalances. With a newly designed Multi-Center Loss, the model is allowed to learn invariant features in each data domain instead of overfitting a certain distribution to solve the multi-domain adaptation problem.

\subsection{Coarse-Grained Leading Forest}

 We designed a new clustering algorithm coarse-grained leading forest in combination with the construction of OLeaF \cite{xu2021lapoleaf}, and its construction process is shown in Algorithm \ref{algo:clf-construction}.

{\color{black}
1) \textbf{Calculate the distance matrix and sample density}: First calculate the distance matrix between the samples in the dataset $\boldsymbol{X}$ using an arbitrary distance metric, and then calculate the density of each sample point $i$ with reference to Eq. (\ref{eq:density}).}

\begin{equation}
\label{eq:density}
\rho_i=\sum_{j\in \boldsymbol{I} \backslash\{i\}\backslash \textbf{O}_i}\exp\big(-(d_{i,j}/d_{rd})^2\big),
\end{equation}
where $\boldsymbol{I}=\left\{1,2,...,N\right\}$ is the index set of dataset $\boldsymbol{X}$, $d_{rd}$ is the radius of the hypersphere for computing density centered at $\boldsymbol x_i$, $\boldsymbol{O}_i$ is the set of samples lie outside of the hypersphere that do not contribute to $\rho_i$, and the symbol ``$\backslash$'' denotes the set difference operation between the two sets.

{\color{black}
2) \textbf{Creating coarse-grained nodes}: First arrange the samples in order of decreasing density values, denoted as $\boldsymbol{S}$ the indices of the sorted data, i.e., $\boldsymbol{S}_i$ is the index of the data point with the $i$-th largest density value. Then iterate through the samples in $\boldsymbol{S}$ sequentially, and if the data point $\boldsymbol{S}_i$ has not yet been merged into a coarse-grained node, then the data point $\boldsymbol{S}_i$ and the points within $d_{rn}$ distance from it are formed into a coarse-grained node, i.e.:}


\begin{equation}
\label{eq:coarse-node}
\boldsymbol{C}(i) = \{\boldsymbol{S}_i\}\cup \boldsymbol{K}(i)\  \backslash\ \boldsymbol{A},\quad \boldsymbol{S}_i\notin \boldsymbol{A},
\end{equation}
where $\boldsymbol{C}(i)$ are members of the newly generated coarse-grained node, $\boldsymbol{S}_i$ will be a prototype for that coarse-grained node, $\boldsymbol{K}(i)$ is the set of nodes that are within $d_{rn}$ from $\boldsymbol{S}_i$, i.e., $\boldsymbol{K}(i) = \{j\ |\ j\in \boldsymbol{I},\ d_{S_i,j}<d_{rn}\}$, and $\boldsymbol{A}$ is the set of visited nodes, i.e., the set of nodes that have already been merged into a particular coarse-grained node. Note that if $\boldsymbol{S}_i$ itself is already in $\boldsymbol{A}$ then it skips the creation of coarse-grained node and continues to process the next node in $\boldsymbol{S}$.

{\color{black}
3) \textbf{Finding the leading node}: Whenever a new coarse-grained node is created, a lead node is found for it. Here, the problem is transformed into finding the leading node of the prototype $\boldsymbol{S}_i$ of a coarse-grained node, denoted as $l_i$, and then $\boldsymbol{C}(l_i)$ is assigned as the leading node of $\boldsymbol{C}(i)$. The process of finding the leading node $\boldsymbol{C}(l_i)$ of $\boldsymbol{C}(i)$ can be formulated as Eq. (\ref{eq:leading-node}):}

\begin{equation}
\label{eq:leading-node}
l_{i}=\arg\min_j\left\{ d_{S_i,j}|\rho_j>\rho_{S_i} \right\}, j\in{\boldsymbol{I}\backslash \{\boldsymbol{S}_i}\}\backslash \boldsymbol{O}_{S_i},
\end{equation}
 where $\boldsymbol{O}_{S_i}$ is the set of nodes whose distance from $\boldsymbol{S}_i$ exceeds $d_{rd}$. Note that $l_{i}$ may not exist, and when $l_{i}$ is not found, the coarse-grained node $\boldsymbol{C}(i)$ is the root of a leading tree in the whole coarse-grained leading forest. Also, since the density of nodes in $\boldsymbol{S}$ is decreasing, when $l_{i}$ is found, it must have been processed and already merged into  $\boldsymbol{C}(i)$ {\color{black} since the density of $l_{i}$ is higher.}

{\color{black}
\textbf{Time Complexity.} The time complexity of calculating the distance matrix between sample points is $O(N^2)$, where $N$ is the number of sample points. Next, calculating the density for each sample point also has a complexity of $O(N^2)$. Following this, the time complexity for sorting the densities in descending order is $O(N \log N)$. Within the main loop, each iteration may need to check all unvisited points to determine which points are within the distance $d_{rn}$, with the worst-case complexity being $O(N)$. Similarly, when finding the node with the highest density within distance $d_{rd}$, the worst-case complexity is also $O(N)$. Overall, the primary time-consuming operations are the distance and density calculations, resulting in an overall time complexity of $O(N^2)$.

\textbf{Space Complexity.} The main cost arises from storing the distance matrix between sample points, with a complexity of $O(N^2)$. Additionally, storing the density information for each sample point, the sorting indices, and the boolean vector all have complexities of $O(N)$. Consequently, the overall space complexity of the algorithm is also $O(N^2)$.

Although the time and space complexity of the algorithm is relatively high, it spends mainly on the computation and storage of the distance matrix, operations that can be optimized by distributed or parallel methods. A number of acceleration schemes for density-peak clustering have been proposed. For example, in another work of ours, FaithPDP \cite{Xu2024Faithful}, which is specifically optimized for the construction of the leading tree structure, a parallel solution is proposed, while instead of storing the entire distance matrix, only two vectors and a tall matrix need to be stored. These improvements significantly improve the efficiency of the algorithm, especially when dealing with large-scale datasets.}

\begin{algorithm}[htbp]
\caption{\label{algo:clf-construction}Construction of CLF}
\SetKwInOut{Param}{Parameter}
\KwIn{All training samples of a given category $\textbf{\textit{X}}_c$.}
\KwOut{A Coarse-Grained Leading Forest $\textit{clf}$ for a given category.}
\Param{$d_{rd}$ as the hypersphere radius to compute density, $d_{rn}$ as the radius of the coarse-grained node.}

\BlankLine
$N$ = number of samples in $\textbf{\textit{X}}_c$\;
\textit{dist} = calculate\_distance($\textbf{\textit{X}}_c$)\;
\textit{density} = calculate\_density(\textit{dist}, $d_{rd}$)\;
$\boldsymbol{S}$ = argsort(\textit{density}, descend=$\textbf{T}$)\;
\tcp{\small Record the index of accessed samples }
$\boldsymbol{A}$ = initial\_vector($N$, $\textbf{F}$)\;

\For{$idx$ \textbf{in} range($N$)}{
    \textit{i} = $\boldsymbol{S}[idx]$\;
    \If{$\boldsymbol{A}$[\textit{i}]==$\textbf{T}$}{
        \textbf{continue}\;
    }
    \tcp{\small Combine unvisited points within $d_{rn}$}
    $\boldsymbol{C}(i)$ = $\{i\} \bigcup \{j|\boldsymbol{A}[j] == $\textbf{F}$ \; \&\&\;  \textit{dist}[\textit{i,j}] \leq d_{rn})\}$\;
    $\boldsymbol{A}[j] $= $\textbf{T}$ for $j \in \boldsymbol{C}(i)$\;
        $l_i$ = find\_leader($i$, $d_{rn}$)\;
    \tcp{\small $\boldsymbol{C}(i)$ becomes a root if no $l_i$ exists}
    \If{$l_i \neq$ null}{
        $\boldsymbol{C}(i)$.\textit{leader} = $\boldsymbol{C}(l_i)$\;
    }
    \Else{
        \textit{clf.root}.append($\boldsymbol{C}(i)$)\;
    }
}

\Return \textit{clf}\;
\end{algorithm}

\subsection{Sampling based on CLF}

\begin{figure*}[htbp]
\centering
\includegraphics[width=0.96\linewidth]{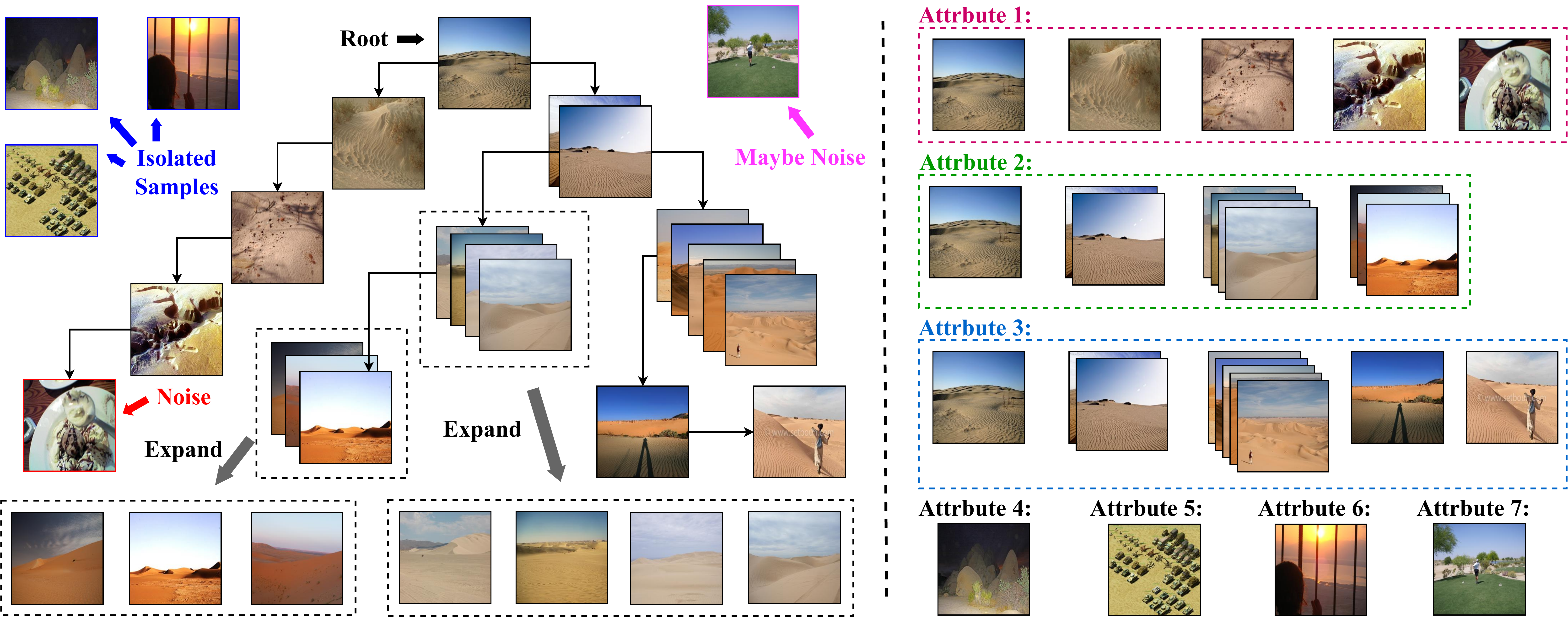}
\caption{The left is an example of CLF constructed for category ``sand'', while the right is an example of attribute splitting using CLF. Each path from the root to the leaf node can be considered as an (implicit) attribute, and the samples within the coarse-grained nodes are very similar that require an appropriate reduction in sampling weights. In addition, the samples within the red and pink boxes demonstrate the potential of CLF for noisy recognition.}
\label{fig:clf_sand_sampling}
\end{figure*}

By constructing the CLF we can portray the attribute distributions within a category, and by adjusting the hyperparameters $d_{rn}$ and $d_{rd}$, we can control the granularity of the distributions portraying. An exemplar CLF in an environment of different attribute distributions is shown in Fig. \ref{fig:clf_sand_sampling}. One can categorize all the data points on each path from root to a leaf node as members of a certain attribute, because {each path represents a new direction of attribute evolution}.



After the samples of each attribute is determined, we can refer to the idea of class-wise resampling \cite{kang2019decoupling} to resample on the attributes (compute the probability of each sampled objects from class $j$ ), {\color{black}as shown in Eq. (\ref{eq:resample}).}
\begin{equation}
\label{eq:resample}
p_j = \frac{(n_j)^q}{\sum_{i=1}^C (n_i)^q},
\end{equation}
where $j$ represents the class index, $n_j$ is the number of samples of class $j$, and $C$ the total number of classes. {\color{black}And when performing intra-class sampling, $j$ represents the attribute index, $n_j$ is the number of samples of attribute $j$, and $C$ the total number of attributes. $q \in [0,1]$ is a user-defined parameter, and balanced sampling is performed when $q = 0$.

\textbf{Remark 2}: \emph{Both inter-class and intra-class resampling refer to Eq. (\ref{eq:resample}), except that the meaning of the symbols in the equations is different for the two types of sampling.}}

As shown in Algorithm \ref{algo:clf-sampling}, when sampling different attributes within a category, the idea is the same as class-wise sampling. When applying Eq. (\ref{eq:resample}), the only difference is to regard $j$ as the attribute index, $n_j$ as the number of samples of attribute $j$, and $C$ as the total number of attributes. Besides, it should be noted that the same sample may belongs to more than one attribute collection (e.g., the root node appears in all branches of the same tree, so it has all the attributes under consideration), so the weight of such a sample needs to be penalized. At the same time, due to the concept of coarse-grained node in our algorithm, as shown in Fig. \ref{fig:clf_sand_sampling}, the members in the coarse-grained nodes are highly alike, so the sampling weight of the members in the coarse-grained nodes should be reduced accordingly. In this paper, we let the members of a coarse node $\boldsymbol{C}(i)$ equally share the weight of $\boldsymbol{C}(i)$.

We provide a detailed description on attribute balanced sampling based on the CLF in Fig. \ref{fig:clf_sand_sampling} (ignoring isolated samples for demonstration purposes). Firstly, each path from root to a leaf node reflects the evolution along an attribute, which means that if you want to evenly sample the data of each attribute, you only need to assign equal sampling weights to the collection of data in each path. Where for nodes that are repeated in multiple paths, we simply divide their sampling weight by the number of repetitions as a penalty. Take the example of calculating the weight of the root node, firstly since there are three attribute groups, so $weight = \frac{1}{3}$, and then the node's weight in the three attribute groups are $\frac{1}{5}$, $\frac{1}{4}$, and $\frac{1}{5}$, so the root node's sampling weights globally are $\frac{1}{15}$, $\frac{1}{12}$, and $\frac{1}{15}$, and summing up the three weights yields $weight = \frac{13}{60}$, and finally an appropriate penalty is to be applied, i.e., $weight = \frac{weight}{repetition} = \frac{13}{180}$. Furthermore, for a CoarseNode containing multiple samples, the penalty is similar, setting the sampling weight of each sample to $weight = CoarseNode.weight/CoarseNode.length$. Taking the second coarse-grained node in Attribute 2 as an example, the sampling weight of this node is $weight = (\frac{1}{3}\times\frac{1}{4} + \frac{1}{3}\times\frac{1}{5})/2 = \frac{3}{40}$, and the sampling weight of each sample in this coarse-grained node is $weight = \frac{3/40}{2} = \frac{3}{80}$.

{\color{black}
\textbf{Remark 3}: \emph{The weights calculated for each sample in the above steps are only relative values (i.e., do not necessarily sum to one) and need to be normalized after all sample weights have been calculated.}}

\begin{algorithm}[htbp]
  \caption{\label{algo:clf-sampling} Attribute-wise sampling via CLF}
  \SetKwInput{Input}{Input}
  \SetKwInput{Output}{Output}
  \Input{A Coarse-Grained Leading Forest \textit{clf}, balance factor \textit{q}}
  \Output{The sampling probability of each sample within a certain category weights}
  \BlankLine
  \textit{weights} = initial\_vector(default=0, size=$|$\textit{clf}$|$)\;
  \textit{paths} = generate\_path(\textit{clf})\;
  \textit{repetitions} = get\_repetition(\textit{paths})\;
  \textit{path\_weight} = get\_path\_weight(\textit{paths}, \textit{q}) using Eq. (\ref{eq:resample})\;
  \ForEach{\textit{path} \textbf{in} \textit{paths}}{

    \textit{node\_weight} = \textit{path\_weight} / $\text{len}(\textit{path})$\;
    \tcp{\small All nodes are considered coarse }
    \ForEach{\textit{node} \textbf{in} \textit{path}}{
      \textit{node\_weight} /=  $\text{len}(\textit{node.members})$\;
          }
  }
  \textit{weights} $/= \textit{repetitions}$\;
  {\color{black}\textit{weights} $/= \text{sum}(\textit{weights})$}\;
  \Return \textit{weights}\;
\end{algorithm}

\subsection{Multi-Center Loss}

\begin{figure*}[htp]
\centering
\includegraphics[width=0.9\linewidth]{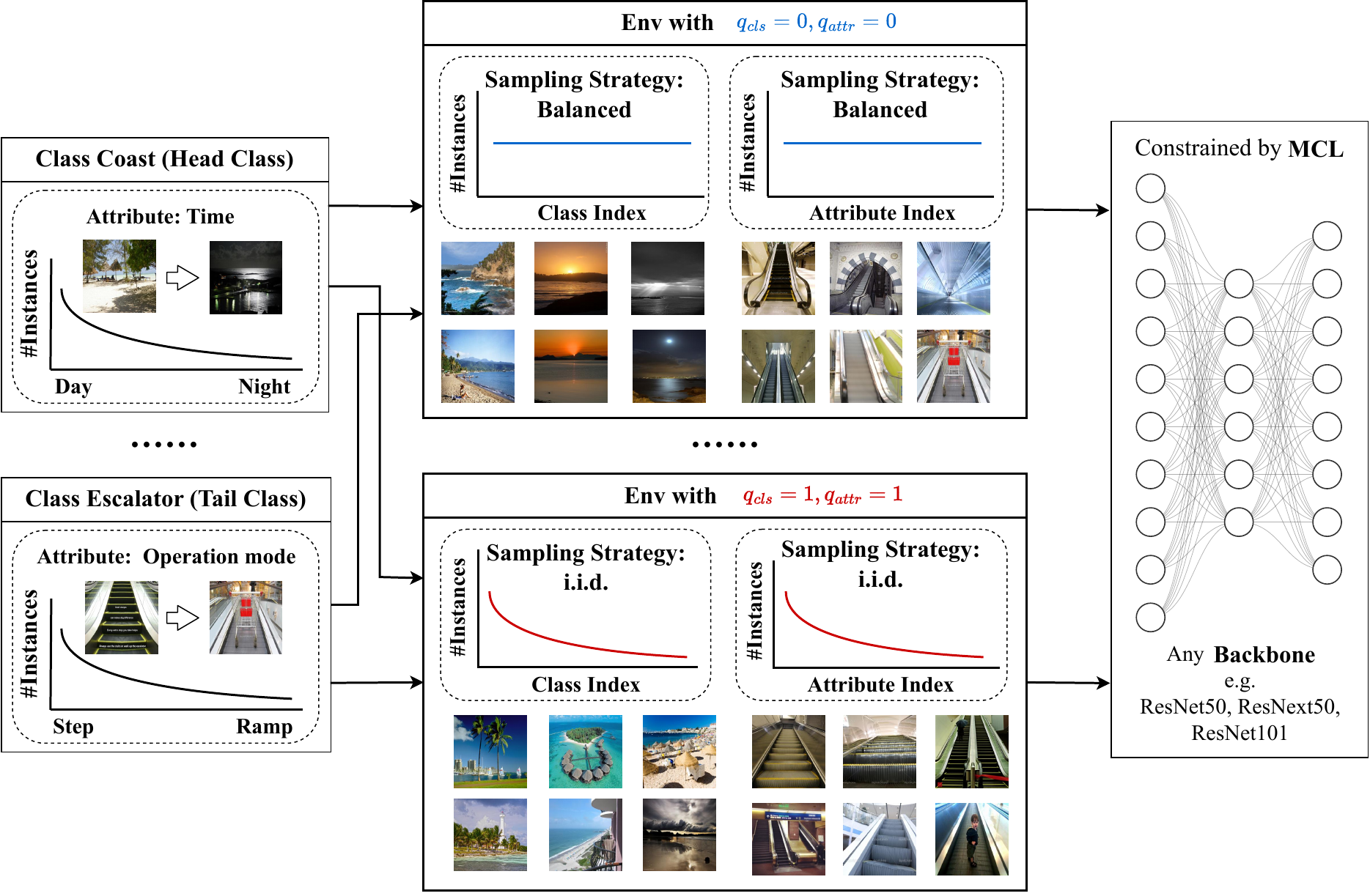}
\caption{Overall framework diagram. Different environments have different sampling strategies, where $q_{cls}$ and $q_{attr}$ are balancing factors for class-wise sampling and attribute-wise sampling, respectively.}
\label{fig:framework}
\end{figure*}

Multiple training environments for invariant feature learning were constructed in the previous step, and the next might be training the model with the objective function of IRM \cite{arjovsky2020invariant}. However, since the original IRM loss does not converge occasionally in experiments, we designed a new goal Multi-Center Loss based on the idea of IRM and the center loss in IFL \cite{tang2022invariant}, which can be formulated as:
\begin{equation}
\label{eq:mcl1}
\begin{split}
&\mathop{\min}\limits_{\theta,\ w}\sum\limits_{e\in \mathcal{E}}\sum\limits_{i}L_{cls}\big(f(x_i^e;\ \theta),\ y_i^e;\ w\big)\\
\mathrm{ s.t. }&\quad \theta\in\mathop{\arg\min}\limits_{\Theta}\sum\limits_{e\in\mathcal{E}}\sum\limits_{i}{\Vert f(x_i^e;\ \Theta)-\mathcal{C}(x_i)\Vert}_2,
\end{split}
\end{equation}
where $\Theta$ and $w$ are the learnable parameters of the backbone and classifier, respectively, $x_i^e$ and $y_i^e$ are the $i$-th instance in environment $e$ and its label, respectively, $\mathcal{E}$ is all the training environments, $f(x_i^e;\ \theta)$ is the feature extracted by the backbone from $x_i^e$, $L_{cls}\big(f(x_i^e;\ \theta),\ y_i^e;\ w\big)$ is the classification loss under environment $e$ (with arbitrary loss function), and $\mathcal{C}(x_i)$ is the learned feature of the center to which $x_i^e$ belongs (in all environments $\mathcal{E}$). Note that the number of centers in each category $n_{c_{y_i}}\geq1$, and the center of $x_i^e$ depends on which tree in the CLF contains $x_i^e$, i.e., $n_{c_{y_i}} \equiv n_{t_{y_i}}$, where $n_{t_{y_i}}$ is the number of trees in the CLF constructed from all samples of that category, and the initial value of $\mathcal{C}(x_i)$ is the value of the \textit{prototype} (the root) of the tree containing $x_i^e$. The practical version of this optimization problem is shown in Eq. \ref{eq:mcl2}, where $L_{IFL}={\Vert f(x_i^e;\ \theta)-\mathcal{C}(x_i)\Vert}_2$ is the constraint loss for invariant feature learning and $\alpha >0$ is a trade-off parameter:
\begin{equation}
\label{eq:mcl2}
\begin{split}
\mathop{\min}\limits_{\theta,\ w}\sum\limits_{e\in\mathcal{E}}\sum\limits_{i}L_{mc}
&=\mathop{\min}\limits_{\theta,\ w}\sum\limits_{e\in\mathcal{E}}\sum\limits_{i} L_{cls} + \alpha\cdot L_{IFL} \\
&=\mathop{\min}\limits_{\theta,\ w}\sum\limits_{e\in\mathcal{E}}\sum\limits_{i}L_{cls}\big(f(x_i^e;\ \theta),\ y_i^e;\ w\big) \\
&\quad \ + \alpha\cdot{\Vert f(x_i^e;\ \theta)-\mathcal{C}(x_i)\Vert}_2.
\end{split}
\end{equation}

\begin{figure}[htbp]
\centering
\includegraphics[width=1.0\linewidth]{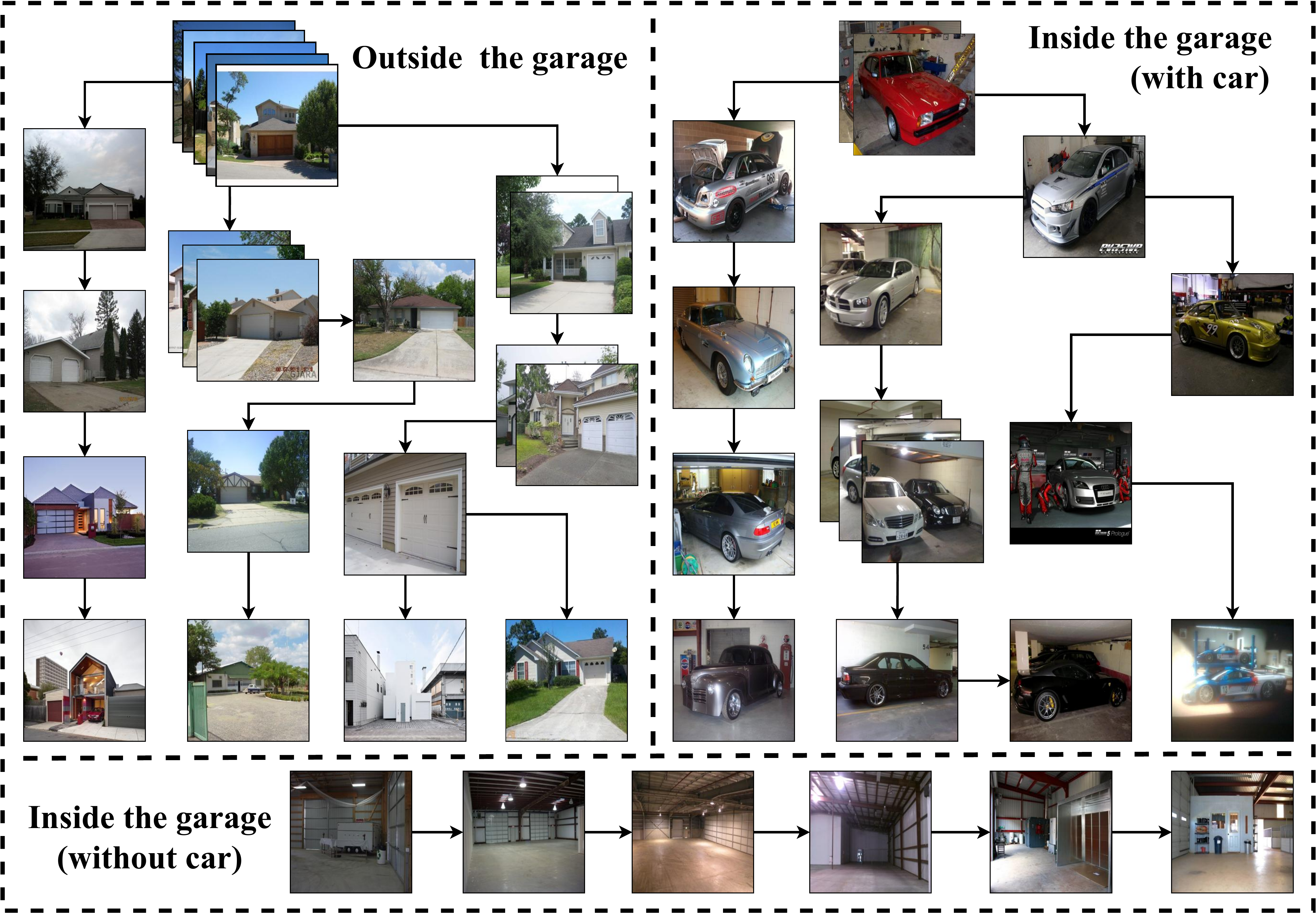}
\caption{Multiple trees in the CLF of category ``garage" (only part of it is shown), there are multiple subclasses in the category ``garage", e.g., outside the garage, inside the garage (parked cars), and inside the garage (no parked cars), and there is a huge disparity among these subclasses.}
\label{fig:clf_gap}
\end{figure}

This loss is the IRM version of center loss that improves the robustness over the original version of center loss. As in the previous introduction of CLF, for some artificial datasets, even within the same category there may be great the difference between samples, i.e., the number of trees in CLF is greater than one, as shown in Fig. \ref{fig:clf_gap}. The category ``garage" can actually be divided into three categories: ``Outside the garage", ``Inside the garage with car", and ``Inside the garage without car". The features of these three subcategories vary greatly. Using only one center would make each category's features gradually approach one center during the training process, which will actually degrade the quality of learned features. This is the exact motivation for proposing Multi-Center Loss.

\subsection{\textsc{Cognisance} Framework}
The overall framework of \textsc{Cognisance} is shown in Fig. \ref{fig:framework}. Each environment is sampled according to a pair of parameter $(q_{cls},q_{attr})$, with $(q_{cls}$ being the balance factor for class-wise sampling and $(q_{attr}$ being the balance factor for attribute-wise sampling. The model parameters can be shared for training under the constraint of MCL. The paths to portray the implicit attribute distribution during sampling is achieved through CLF, while the number of centers in MCL is determined by the number of trees in CLF of the corresponding category.

The algorithmic description of \textsc{Cognisance} is shown in Algorithm \ref{algo:overall}, which consists of two stages:

1) Since the initial features of the samples need to be used for clustering when constructing CLF, $M$-round normal sampling training is required to obtain an initial model with imperfect predictions;

2) The initial features are used to construct the CLF, and different environments are constructed through the CLF and different balance factor pairs.
For example, in the experimental phase, this article sets up two environments with balance factor pairs of $(q_{cls}^{e_2}=0, q_{attr}^{e_2}=0)$ and $(q_{cls}^{e_1}=1, q_{attr}^{e_1}=1)$, where the former is a balanced sampling environment for both categories and attributes and the latter is a normal i.i.d. sampling environment that exhibits both class-wise and attribute-wise imbalance. Then, the feature learner is continuously updated and the centers in the CLF and MCL are updated accordingly. The number of epoch steps for executing updates in the second stage can be adjusted, rather than being fixed to one update per epoch.

\begin{algorithm}[htbp]
\caption{\textsc{Cognisance} framework}
\label{algo:overall}
\KwIn{An original training set $\{(x, y)\}$; balance parameter pairs $\{(q_{cls}^e, q_{attr}^e)\}$ for different environments.}
\KwOut{backbone $f(\cdot; \theta)$, classifier $g(\cdot; w)$}
\BlankLine
Initialize: backbone $f(\cdot; \theta)$, classifier $g(\cdot; w)$\;
\For{$M$ warm-up epochs}{
    \tcp{\small Optimize the model for any loss}
    $\theta, w \leftarrow \hat{\theta}, \hat{w} \in \text{arg min}_{\theta, w} L_{cls}(g(f(x; \theta);w), y)$\;
}
$clf = \text{ClfConstruct}(\{(x, y)\}, \theta)$ (Alg. \ref{algo:clf-construction})\;
$\mathcal{E} = \text{EnvConstruct}(\{(q_{cls}^e, q_{attr}^e)\}, clf )$ (Alg. \ref{algo:clf-sampling})\;
\For{$N$ epochs}{
    $\{(x^{e}, y^{e})\} = \text{DataLoader}(\mathcal{E})$\;
    $\{C_y\} = \text{ReadCenters}(clf )$\;
    $\theta, w \leftarrow \hat{\theta}, \hat{w} \in \text{arg min}_{\theta, w} L_{cls}(g(f(x; \theta);w), y) + \alpha \cdot L_{IFL}(f(x; \theta), \{C_y\})$\;
    $clf = \text{ClfConstruct}(\{(x, y)\}, \theta)$\;
    $\mathcal{E}= \text{EnvConstruct}(\{(q_{cls}^e, q_{attr}^e)\}, clf )$\;
}
\Return backbone $f(\cdot; \theta)$, classifier $g(\cdot; w)$\;
\end{algorithm}

{\color{black}
\subsection{Noise Selection}

Identifying noise in long-tailed datasets is a highly challenging task \cite{lu2023label}. Due to the low frequency of tail samples in long-tailed datasets, accurately distinguishing genuine tail categories from noise becomes extremely difficult. In such cases, the number of noise samples may be comparable to or even exceed that of tail category samples, thus increasing the complexity of model training and application. Especially, when using resampling methods, the resampling process may lead to an increase in the proportion of noise samples in the dataset, thereby negatively impacting model training.

\begin{figure}[htbp]
    \centering
    \includegraphics[width=1.0\linewidth]{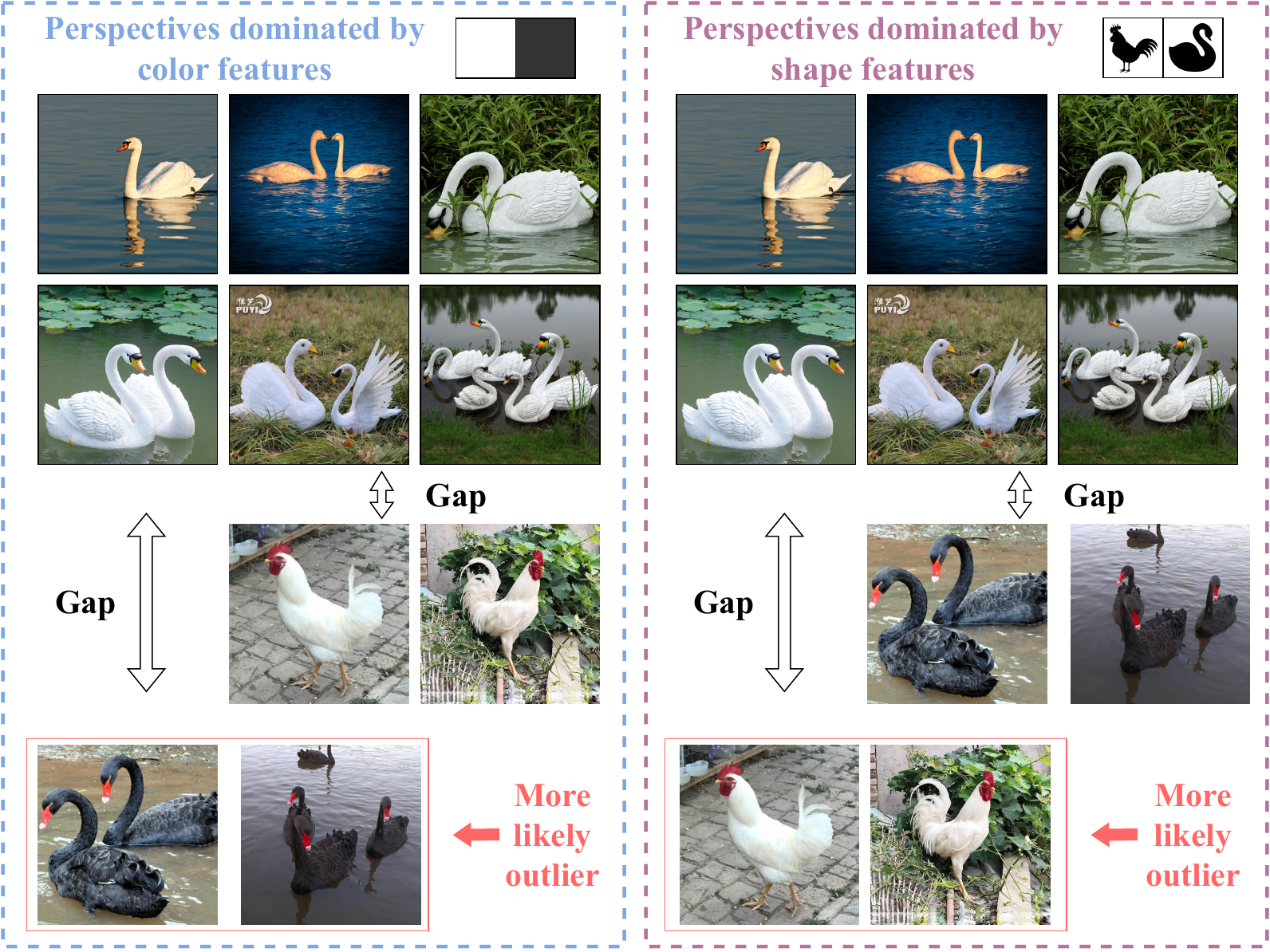}
    \caption{Distribution of noise samples in clustering under different feature dominances; noise samples are more peripheral under invariant features.}
    \label{fig:inv_attr2}
\end{figure}

To tackle this challenge, we propose a noise selection algorithm based on the \textsc{Cognisance}. The algorithm has two main motivations: First, by learning invariant features from the data, normal and noise samples can be clearly distinguished in the feature space. As shown in Fig. \ref{fig:inv_attr2}, when clustering based on pseudo-features, the ``black swan'' appears to be very outlier due to its rarity. If outlier-ness is simply used as the standard for identifying noise samples, the ``black swan'' is likely to be misidentified as noise, while the ``white rooster'' might be misidentified as a normal sample. However, by removing the pseudo-feature ``color'' and instead using the invariant feature ``shape'' as the main basis for identification, the ``black swan'' and ``white swan'' will cluster together, while the ``white rooster'' will appear as an outlier due to the shape difference.

Next, based on visualizations from previous work, we observed the distribution characteristics of noise samples in guiding forests. As shown in the red boxes of Fig. \ref{fig:evolution1} and Fig. \ref{fig:clf_sand_sampling}, noise samples are often located at the bottom of leading trees and, due to their outlier nature, the number of nodes forming a tree is usually sparse. Based on these \textbf{quantity and location characteristics}, we designed the core noise selection strategy:

1) \textbf{Cluster Size-Based Noise Labeling}: For small clusters of samples, if the number of samples in the cluster is less than the threshold \( N_{\text{min}} \), all samples in the cluster will be directly labeled as noise. This criterion is designed based on the characteristic that noise samples typically form smaller clusters in the coarse-grained leading tree.

2) \textbf{Depth and Layer-Based Noise Selection}: In addition, two parameters \( N_{d} \) and \( N_{l} \) are set, representing the depth at which noise selection starts in each coarse-grained leading tree and the number of layers selected from bottom to top, respectively. For example, when \( N_{l}=1 \), all samples in the last layer of the coarse-grained leading tree are marked as noise. The setting of \( N_{d} \) is to reduce the possibility of mislabeling normal samples as noise, as the deeper the sample is in the tree, the higher its ``outlierness'', and the more likely it is to be a noisy sample. This criterion is based on the characteristic that noise samples are often located at deeper levels in the coarse-grained leading tree.

3) \textbf{Density Percentile-Based Noise Filtering}: Finally, to reduce the risk of selecting normal samples by mistake, this scheme introduces a filtering mechanism based on density percentiles at the end of the process. Density to some extent represents outliers, and by screening samples with lower density, the possibility of mislabeling normal samples can be reduced. Setting the density percentile threshold \( P_{d} \) can limit the maximum number of labeled noise samples. For example, when \( P_{d}=10\% \), the number of selected noise samples will not exceed 10\% of the total number of samples in that category, and these noise samples are the 10\% with the lowest density value among all samples. This step enhances the controllability of noise selection, preventing too many samples from being marked as noise.

To label a sample as noise, it is sufficient to satisfy either of the first two criteria (cluster size-based or depth and layer-based) and simultaneously satisfy the third criterion (density percentile-based).

To better highlight the quantity and location characteristics of noise samples, this paper introduces a new metric learning loss, the Multi-Center Triplet Loss (MCTL). This loss function not only brings samples of the same class closer to the center, but also pushes samples of different classes further apart, as shown in Eq. (\ref{eq:mctl2}):
\begin{equation}
\label{eq:mctl2}
\begin{split}
	&\mathop{\min}\limits_{\theta,\ w}\sum\limits_{e\in\mathcal{E}}\sum\limits_{i}L_{mc}
=\mathop{\min}\limits_{\theta,\ w}\sum\limits_{e\in\mathcal{E}}\sum\limits_{i} L_{cls} + \alpha\cdot L_{IFL} \\
&=\mathop{\min}\limits_{\theta,\ w}\sum\limits_{e\in\mathcal{E}}\sum\limits_{i} L_{cls}(g(f(x_i^e;\ \theta);w)\ ,y_i^e) \\
&\quad + \alpha\cdot\mathop{\max}(0,{\Vert f(x_i^e;\ \hat{\theta})-C_p(x_i^e)\Vert}_2 \\
&\quad - {\Vert f(x_i^e;\ \hat{\theta})-C_n(x_i^e)\Vert}_2),
\end{split}
\end{equation}
where $L_{IFL}$ is the invariant feature learning constraint loss, and $\alpha$ is the balancing parameter. This loss function is an IRM version of the triplet loss, which further enhances the robustness of the algorithm compared to the multi-center loss. As shown in Fig. \ref{fig:noise_clf}, it is precisely because MCTL widens the distance between samples of different categories that the position of noisy samples can be closer to the bottom.

\begin{figure}[htbp]
    \centering
    \includegraphics[width=1.0\linewidth]{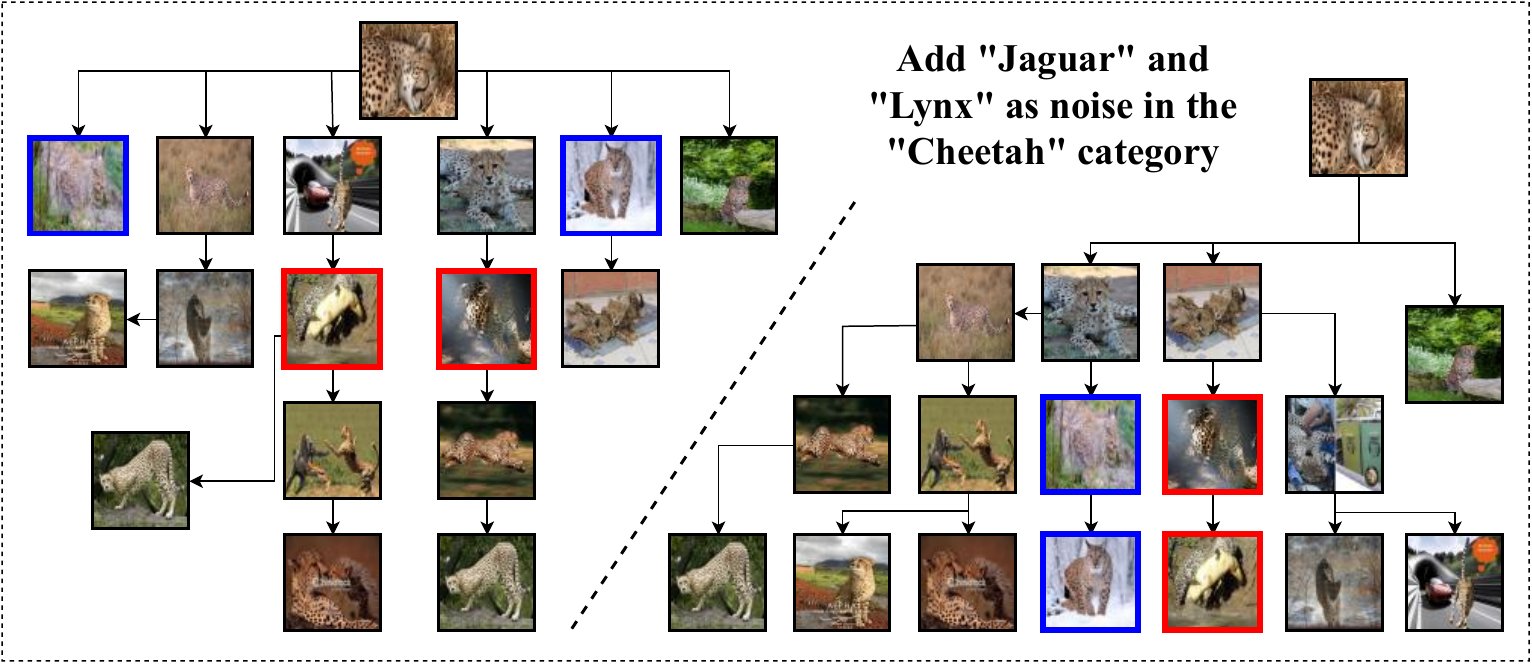}
    \caption{{\color{black}The CLF constructed based on the learned representations using different loss functions. The left figure shows results using ordinary cross entropy loss, while the right figure depicts results from MCTL. The red and blue boxes represent noise from ``Jaguar'' and ``Lynx'', respectively. In the right, the noise samples are concentrated at the bottom.}}
    \label{fig:noise_clf}
\end{figure}

We refer to this algorithm designed specifically for long-tailed noise datasets as the \textsc{Cognisance}$^+$ framework. The specific process is shown in algorithm \ref {algo:overall2}. Compared with the \textsc{Cognisance} framework, this method can further improve model performance on noisy datasets.

\begin{algorithm}[htbp]
\caption{Overall Framework Training Process}
\label{algo:overall2}
    \KwIn{An original training set $\{(x, y)\}$; balance parameter pairs $\{(q_{cls}^e, q_{attr}^e)\}$ for different environments.}
    \KwOut{backbone $f(\cdot; \theta)$, classifier $g(\cdot; w)$}
    \BlankLine
    Initialize: backbone $f(\cdot; \theta)$, classifier $g(\cdot; w)$\;
    \For{$M$ warm-up epochs}{
        \tcp{\small Optimize the model for any loss}
        $\theta, w \leftarrow \hat{\theta}, \hat{w} \in \text{arg min}_{\theta, w} L_{cls}(g(f(x; \theta);w), y)$\;
    }
    \textit{noises} = \textit{initial\_list}()\;
    $\{F_y\} = \textit{construct\_clf}(\{(x, y)\}, \theta, w)$\;
    $\{e_n\} = \textit{construct\_env}(\{(q_{cls}^e, q_{attr}^e)\}, \{F_y\}, \textit{noises})$\;
    \For{$N$ epochs}{
        $\{(x^{e}, y^{e})\} = \textit{data\_load}(\{e_n\})$\;
        $\{(C_p^x, C_n^x)\} = \textit{read\_centers}(\{F_y\}, \{(x^{e}, y^{e})\})$\;
        $\theta, w \leftarrow \hat{\theta}, \hat{w} \in \text{argmin}_{\theta, w} L_{cls}(g(f(x; \theta);w), y) + \alpha \cdot L_{IFL}(f(x; \theta), \{(C_p^x, C_n^x)\})$\;
        $\{F_y\} = \textit{construct\_clf}(\{(x, y)\}, \theta)$\;
        \tcp{Iteratively update noise samples and environment.}
        $\textit{noises} = \textit{select\_noises}(\{F_y\})$\;
        $\{e_n\} = \textit{construct\_env}(\{(q_{cls}^e, q_{attr}^e)\}, \{F_y\}, \textit{noises})$\;
    }
    \Return $f(\cdot; \theta)$, $g(\cdot; w)$
\end{algorithm}
}

\section{Experiment}\label{sec:Experiment}

\subsection{Evaluation Protocols}

Before conducting the experiment, it is necessary to first introduce two new evaluation protocols: CLT Protocol and GLT Protocol, both of which were proposed in the first baseline of GLT \cite{tang2022invariant}:
\begin{itemize}
\item \textbf{Class wise Long Tail (CLT) Protocol}: The samples in the training set follow a long-tailed distribution, which means that they are normally sampled from the LT dataset, while the samples in the test set are balanced by class. Note that the issue of attribute distribution is not considered in CLT, as the training and testing sets of CLT have the same attribute distribution and different class distributions, so the effectiveness of category long-tailed classification can be evaluated.
\item \textbf{Generalized Long Tail (GLT) Protocol}: Compared with CLT protocol, the difference in attribute distribution is taken into account, that is, the attribute bias in Eq. (\ref{eq:glt}) is introduced. The training set in GLT is the same as that in CLT and conforms to the LT distribution, while the attribute distribution in the test set tends to be balanced. Since attribute-wise imbalance is ubiquitous, the training set and test set in GLT have different attribute distributions and different class distributions. Therefore, it is reasonable to evaluate the model's ability to handle both class-wise long-tailed and attribute-wise long-tailed classification with GLT protocol.
\end{itemize}



\subsection{Datasets and Metrics}

\newcolumntype{C}{c!{\vline width 1pt}}
\begin{table*}[htbp]
\centering
\renewcommand{\arraystretch}{1.5}
\caption{Evaluation of CLT and GLT protocols on ImageNet-GLT. The family of ?+\textsc{Cognisance} won for 43 times out of 48 comparisons against the competing models.}
\label{tab:result_imgnet}

\resizebox{\textwidth}{!}{
\begin{tabular}{c|Cc|c|c|Cc|c|c|c}
\hline
\multicolumn{2}{C}{\textbf{Protocols}} & \multicolumn{4}{C}{\textbf{Class-Wise Long Tail (CLT) Protocol}} & \multicolumn{4}{c}{\textbf{Generalized Long Tail (GLT) Protocol}} \\ \hline
\multicolumn{2}{C}{$\langle$\textbf{Accuracy} $|$ \textbf{Precision}$\rangle$}
& $Many_C$ & $Medium_C$ & $Few_C$ & $\textbf{Overall}$
& $Many_C$ & $Medium_C$ & $Few_C$ & $\textbf{Overall}$ \\\hline
\multirow{8}{*}{\rotatebox[origin=c]{90}{\textbf{Re-balance}}}
& \multicolumn{1}{C}{Baseline}           & $58.39|38.35$ & $36.00|52.15$ & $13.98|55.34$ & $41.65|47.11$ & $50.70|32.50$ & $27.80|43.99$ & $10.18|47.70$ & $34.32|39.95$ \\
& \multicolumn{1}{C}{BBN}                & $62.66|43.45$ & $44.36|55.44$ & $14.66|57.57$ & $47.22|50.96$ & $53.37|36.45$ & $34.97|46.81$ & $10.73|46.08$ & $38.70|42.56$ \\
& \multicolumn{1}{C}{BLSoftmax}          & $54.03|47.07$ & $41.65|46.83$ & $28.37|37.58$ & $44.61|45.54$ & $46.45|40.45$ & $32.67|38.59$ & $21.38|29.57$ & $36.49|37.98$ \\
& \multicolumn{1}{C}{Logit-Adj}          & $53.30|49.05$ & $43.49|44.12$ & $31.86|35.00$   & $45.67|44.72$ & $45.43|41.75$ & $34.20|35.90$ & $24.58|27.78$ & $37.25|37.02$ \\
& \multicolumn{1}{C}{GLTv1}              & $62.51|42.32$ & $38.69|57.23$ & $17.41|\textbf{65.00}$   & $45.03|52.43$ & $54.40|36.23$ & $30.18|49.61$ & $12.64|55.47$ & $37.24|45.14$ \\

\cdashline{2-10}[0.8pt/3pt]
& \multicolumn{1}{C}{* Baseline + \textsc{Cognisance}}  & $\textbf{62.84}|42.82$ & $39.62|57.25$ & $18.61|61.55$ & $45.76|52.12$ & $\textbf{54.53}|36.47$ & $31.3|49.99$ & $13.80|\textbf{58.27}$ & $37.97|45.82$ \\
& \multicolumn{1}{C}{* BLSoftmax + \textsc{Cognisance}} & $58.37|53.88$ & $\textbf{44.98}|51.97$ & $33.82|39.05$ & $\textbf{48.66}|50.80$  & $49.92|46.89$ & $\textbf{36.11}|44.31$ & $26.14|30.03$ & $\textbf{40.14}|43.20$ \\
& \multicolumn{1}{C}{* Logit-Adj + \textsc{Cognisance}} & $43.30|\textbf{75.13}$ & $41.19|\textbf{59.08}$ & $\textbf{45.43}|27.11$ & $42.92|\textbf{60.70}$  & $36.01|\textbf{70.18}$ & $32.60|\textbf{52.48}$ & $\textbf{36.26}|21.76$ & $34.51|\textbf{54.95}$ \\
\specialrule{0.6pt}{0pt}{0pt}
\multirow{4}{*}{\rotatebox[origin=c]{90}{\textbf{Augment}}}
& \multicolumn{1}{C}{Mixup}              & $60.14|38.02$ & $31.46|56.67$ & $\ 7.59|32.82$ & $39.35|45.63$ & $51.68|32.21$ & $23.87|48.25$ & $\  5.47|28.27$ & $32.23|38.84$ \\
& \multicolumn{1}{C}{RandAug}            & $64.14|42.23$ & $40.10|58.27$  & $14.96|59.51$ & $45.94|52.04$ & $55.70|35.87$ & $31.61|50.15$ & $10.20|47.29$ & $38.03|44.01$ \\
\cdashline{2-10}[0.8pt/3pt]
& \multicolumn{1}{C}{* Mixup + \textsc{Cognisance}}     & $67.86|47.90$  & $45.50|62.76$  & $24.98|\textbf{67.56}$ & $51.37|57.54$ & $59.28|40.95$ & $36.16|54.09$ & $17.63|57.72$ & $42.63|49.38$ \\
& \multicolumn{1}{C}{* RandAug + \textsc{Cognisance}}   & $\textbf{69.12}|\textbf{49.31}$ & $\textbf{47.64}|\textbf{63.08}$ & $\textbf{26.97}|67.01$ & $\textbf{53.13}|\textbf{58.16}$ & $\textbf{60.85}|\textbf{42.30}$ & $\textbf{38.46}|\textbf{55.21}$ & $\textbf{19.90}|\textbf{60.11}$ & $\textbf{44.63}|\textbf{50.78}$ \\
\specialrule{0.6pt}{0pt}{0pt}
\multirow{4}{*}{\rotatebox[origin=c]{90}{\textbf{Ensemble}}}
& \multicolumn{1}{C}{TADE}               & $57.30|\textbf{55.22}$  & $46.85|50.29$ & $\textbf{34.69}|37.93$ & $49.21|50.41$ & $49.61|\textbf{48.19}$ & $37.55|42.59$ & $\textbf{27.52}|32.21$ & $40.87|43.28$ \\
& \multicolumn{1}{C}{RIDE}               & $63.18|51.44$ & $47.67|52.55$ & $29.91|47.38$ & $51.21|51.33$ & $54.83|44.02$ & $38.25|44.20$ & $22.77|38.12$ & $42.56|43.21$ \\
\cdashline{2-10}[0.8pt/3pt]
& \multicolumn{1}{C}{* TADE + \textsc{Cognisance}}      & $60.69|55.15$ & $48.15|52.22$ & $33.38|42.12$ & $50.95|51.88$ & $52.77|48.15$ & $39.09|44.06$ & $26.51|33.32$ & $42.68|44.08$ \\
& \multicolumn{1}{C}{* RIDE + \textsc{Cognisance}}      & $\textbf{64.83}|54.60$  & $\textbf{50.95}|\textbf{56.21}$ & $33.28|\textbf{50.15}$ & $\textbf{53.85}|\textbf{54.66}$ & $\textbf{57.10}|47.63$ & $\textbf{42.00}|\textbf{48.42}$ & $25.50|\textbf{41.23}$ & $\textbf{45.57}|{47.03}$ \\
\hline
\end{tabular}}
\end{table*}

\begin{table*}[htbp]
\centering
\renewcommand{\arraystretch}{1.5}
\caption{Evaluation of CLT and GLT protocols on MSCOCO-GLT. The family of ?+\textsc{Cognisance} won for 44 times out of 48 comparisons against the competing models.}
\label{tab:result_mscoco}

\resizebox{\textwidth}{!}{
\begin{tabular}{c|Cc|c|c|Cc|c|c|c}
\hline
\multicolumn{2}{C}{\textbf{Protocols}} & \multicolumn{4}{C}{\textbf{Class-Wise Long Tail (CLT) Protocol}} & \multicolumn{4}{c}{\textbf{Generalized Long Tail (GLT) Protocol}} \\ \hline
\multicolumn{2}{C}{$\langle$\textbf{Accuracy} $|$ \textbf{Precision}$\rangle$}
& $Many_C$ & $Medium_C$ & $Few_C$ & $\textbf{Overall}$
& $Many_C$ & $Medium_C$ & $Few_C$ & $\textbf{Overall}$ \\\hline
\multirow{8}{*}{\rotatebox[origin=c]{90}{\textbf{Re-balance}}}
& \multicolumn{1}{C}{Baseline}           & $81.27|71.08$ & $74.13|76.06$  & $50.17|85.61$ & $71.88|76.15$ & $74.59|64.82$ & $66.25|69.08$ & $35.75|77.22$ & $63.10|69.14$ \\
& \multicolumn{1}{C}{BBN}                & $83.59|70.21$ & $76.13|76.30$ & $47.25|\textbf{90.98}$ & $72.98|77.02$ & $75.36|61.97$ & $68.29|68.61$ & $31.75|\textbf{81.38}$ & $63.41|68.73$ \\
& \multicolumn{1}{C}{BLSoftmax}          & $80.77|71.87$ & $75.67|69.94$ & $45.25|90.08$ & $71.31|74.84$ & $73.23|65.13$ & $68.63|62.64$ & $32.08|79.60$ & $62.81|67.09$ \\
& \multicolumn{1}{C}{Logit-Adj}          & $81.55|73.95$  & $76.00|75.55$ & $\textbf{60.83}|82.04$   & $74.97|76.28$ & $73.64|66.98$ & $68.71|67.61$ & $\textbf{46.58}|70.69$ & $66.00|68.01$ \\
& \multicolumn{1}{C}{GLTv1}              & $82.45|73.09$ & $76.42|79.53$ & $55.58|86.87$   & $74.40|78.60$ & $76.41|67.07$ & $67.33|70.54$ & $39.75|81.02$ & $65.07|71.39$ \\
\cdashline{2-10}[0.8pt/3pt]
& \multicolumn{1}{C}{* Baseline + \textsc{Cognisance}}  & $\textbf{83.00}|73.95$ & $77.88|\textbf{80.48}$ & $55.92|90.08$ & $75.28|\textbf{79.99}$ & $\textbf{76.91}|68.24$ & $67.50|\textbf{71.41}$ & $39.67|80.33$ & $65.31|\textbf{72.05}$ \\
& \multicolumn{1}{C}{* BLSoftmax + \textsc{Cognisance}} & $83.00|74.17$ & $78.37|76.24$ & $53.92|86.51$ & $75.07|77.58$  & $75.86|67.98$ & $70.46|68.14$ & $40.08|80.26$ & $66.22|70.59$ \\
& \multicolumn{1}{C}{* Logit-Adj + \textsc{Cognisance}} & $82.68|\textbf{77.16}$ & $\textbf{80.17}|75.45$ & $59.17|83.54$ & $\textbf{76.78}|77.77$  & $75.14|\textbf{70.56}$ & $\textbf{71.54}|66.34$ & $45.83|73.90$ & $\textbf{67.59}|69.50$ \\
\specialrule{1pt}{0pt}{0pt}
\multirow{4}{*}{\rotatebox[origin=c]{90}{\textbf{Augment}}}
& \multicolumn{1}{C}{Mixup}              & $82.41|72.95$ & $73.12|79.53$ & $54.33|87.47$ & $72.76|78.68$ & $74.64|66.43$ & $65.33|71.00$ & $36.00|78.11$ & $62.79|70.74$ \\
& \multicolumn{1}{C}{RandAug}            & $84.23|74.33$ & $77.42|79.28$  & $56.33|87.07$ & $75.64|79.01$ & $77.91|68.56$ & $69.67|70.95$ & $39.58|78.43$ & $66.57|71.59$ \\
\cdashline{2-10}[0.8pt/3pt]
& \multicolumn{1}{C}{* Mixup + \textsc{Cognisance}}     & $85.00|76.96$  & $81.42|83.19$  & $\textbf{63.33}|88.89$ & $79.03|82.01$ & $79.23|\textbf{70.89}$ & $72.25|\textbf{74.84}$ & $\textbf{46.50}|80.62$ & $\textbf{69.57}|\textbf{74.54}$ \\
& \multicolumn{1}{C}{* RandAug + \textsc{Cognisance}}   & $\textbf{86.55}|\textbf{78.31}$ & $81.50|82.82$ & $62.42|\textbf{90.29}$ & $\textbf{79.47}|\textbf{82.65}$ & $\textbf{79.05}|69.93$ & $\textbf{72.92}|74.51$ & $44.33|\textbf{81.82}$ & $69.33|74.28$ \\
\specialrule{1pt}{0pt}{0pt}
\multirow{4}{*}{\rotatebox[origin=c]{90}{\textbf{Ensemble}}}
& \multicolumn{1}{C}{TADE}               & $83.55|76.47$  & $80.79|73.89$ & $50.50|89.34$ & $75.57|78.06$ & $77.18|68.62$ & $71.58|65.63$ & $33.50|82.32$ & $65.83|70.22$ \\
& \multicolumn{1}{C}{RIDE}               & $84.23|77.56$ & $82.29|77.79$ & $58.42|89.69$ & $78.09|80.16$ & $77.45|71.20$ & $75.17|68.67$ & $41.25|83.33$ & $69.02|72.66$ \\
\cdashline{2-10}[0.8pt/3pt]
& \multicolumn{1}{C}{* TADE + \textsc{Cognisance}}      & $85.59|77.48$  & $82.25|77.85$ & $55.08|90.40$ & $77.90|80.31$ & $79.00|70.54$ & $73.96|68.94$ & $40.67|\textbf{83.65}$ & $68.98|72.59$ \\
& \multicolumn{1}{C}{* RIDE + \textsc{Cognisance}}      & $\textbf{86.27}|\textbf{79.64}$  & $\textbf{83.13}|\textbf{79.89}$ & $\textbf{63.50}|\textbf{90.41}$ & $\textbf{80.26}|\textbf{81.98}$ & $\textbf{80.09}|\textbf{72.77}$ & $\textbf{75.17}|\textbf{71.45}$ & $\textbf{47.83}|83.03$ & $\textbf{71.38}|\textbf{74.34}$ \\
\hline
\end{tabular}}
\end{table*}

We evaluated \textsc{Cognisance} and compared it against the LT and GLT methods on two benchmarks, MSCOCO-GLT, and ImageNet-GLT, which are proposed in the first baseline of GLT \cite{tang2022invariant}.\\
\textbf{ImageNet-GLT} is a long-tailed subset of ImageNet \cite{russakovsky2015imagenet}, where the training set contains 113k samples of 1k classes, and the number of each class ranges from 570 to 4. The number of test set in both CLT and GLT protocols is 60k, and the test set are divided into three subsets according to the following class frequencies: \#sample \textgreater\ 100 for $Many_C$,  100 $\geq$ \#sample $\geq$ 20 for $Medium_C$, and \#sample \textless 20 for $Few_C$. Note that in constructing the test set for attribute balancing in this dataset, the images in each category were simply clustered into 6 groups using \emph{k}-means, and then 10 images were sampled for each group in each category.\\\newline
\textbf{MSCOCO-GLT} is a long-tailed subset of MSCOCO-Attribute \cite{patterson2016coco}, a dataset explicitly labeled with 196 different attributes, where each object with multiple labels is cropped as a separate image. The train set contains 144k samples from 29 classes, where the number of each class ranges from 61k to 0.3k, and the number of test set is 5.8k in both the CLT and GLT protocols, and the test set are divided into three subsets according to the following category frequencies: $Index_C$ $\leq$ 10 for $Many_C$ , 22 $\geq$ $Index_C$ $>$ 10 for $Medium_C$ and $Index_C$ \textgreater\ 22 for $Few_C$ , where $Index_C$ is the index of the category in ascending order.\\\newline
\textbf{Evaluation Metrics}. In the experiments of this paper, three evaluation metrics are used to assess the performance of each methods: 1) Accuracy $\triangleq\frac{\sum_{i=1}^{N}{\rm TP}_i}{\sum_{i=1}^{N}({\rm TP}_i + {\rm FN}_i)}$, which is also used in traditional long-tailed methods for the Top1-Accuracy or micro-averaged recall. In the above two formulas, $N$ denotes the total number of categories, and ${\rm TP}_i$, ${\rm FP}_i$, and ${\rm FN}_i$ denote the number of true positives, false positives, and false negatives in the $i$th category, respectively; 2) Precision $\triangleq \frac{1}{N}\sum_{i=1}^{N}\left(\frac{{\rm TP}_i}{{\rm TP}_i + {\rm FP}_i}\right)$; 3) the harmonic mean of accuracy and precision: $\text{F1-Score}^* \triangleq \frac{2 \times (\text{Accuracy} \times \text{Precision})}{(\text{Accuracy} + \text{Precision})}$, the reason for introducing this metric is to better reveal the accuracy-precision trade-off problem \cite{tang2022invariant} that has not been focused on in the traditional inter-class long-tailed methods.

\subsection{Comparisons with LT Line-up}

\textsc{Cognisance} handles both the class-wise long-tailed problem and the attribute-wise long-tailed problem by eliminating the false correlation caused by attributes, and can be seamlessly integrated with other LT methods. In the following comparison experiments, the backbone of the ``Baseline" is resnext50, and its loss function is cross-entropy loss. For other methods, we follow the state-of-the-art long-tailed research \cite{zhang2023deep} and \cite{tang2022invariant} to classify current long-tailed methods into three categories: 1) Class Re-balancing, 2) Information Augmentation, and 3) Module Improvement. Two or three effective methods from each of these three categories are taken for comparison and enhancement, among which \textbf{BBN} \cite{zhou2020bbn}, \textbf{BLSoftmax} \cite{ren2020balanced} and \textbf{Logit-Adj} \cite{cao2019learning} are chosen for Class Re-balancing category, \textbf{Mixup} \cite{zhang2017mixup} and \textbf{RandAug} \cite{cubuk2020randaugment} for Information Augmentation, and  \textbf{RIDE} \cite{wang2020long} and \textbf{TADE} \cite{zhang2022self} for Module Improvement.

In addition, we have included the first strong baseline GLTv1 \cite{tang2022invariant} in the GLT domain as a comparison. In Table \ref{tab:result_imgnet} and Table \ref{tab:result_mscoco}, the methods with an asterisk are the ones integrated with our component, and the bold-faced numbers are the optimal results in the category of the method. It is evident that \textsc{Cognisance} achieved promising results in all the classifications, especially when combined with the method RandAug or the method RIDE.

\begin{table*}[htbp]
\scriptsize
\centering
\renewcommand{\arraystretch}{1.2}
\caption{Evaluation of ImageNet-GLT and MSCOCO-GLT on LT test set.  The family of ?+\textsc{Cognisance} won for 16 times out of 18 comparisons against the competing models.}
\label{tab:result_lt}
\resizebox{0.7\textwidth}{!}{
\begin{tabular}{c|CccCccc}
\hline
\multicolumn{2}{C}{\textbf{Benchmarks}} & \multicolumn{3}{C}{\textbf{ImageNet-GLT}} & \multicolumn{3}{c}{\textbf{MSCOCO-GLT}} \\ \hline
\multicolumn{2}{C}{\textbf{Overall Evaluation}}& $Acc$ & $Prec$ & $F1$ & $Acc$ & $Prec$ & $F1$ \\\hline
\multirow{8}{*}{\rotatebox[origin=c]{90}{\textbf{Re-balance}}}
& \multicolumn{1}{C}{Baseline}  & 53.93 & 44.46 & 48.74 & 85.74 & 79.98 & 82.76 \\
& \multicolumn{1}{C}{BBN}       & 58.60 & 48.90 & 53.32 & 84.84 & 78.04 & 81.30 \\
& \multicolumn{1}{C}{BLSoftmax} & 51.73 & 41.97 & 46.34 & 83.69 & 73.81 & 78.44 \\
& \multicolumn{1}{C}{Logit-Adj} & 50.94 & 40.58 & 45.17 & 85.28 & 73.60 & 79.01 \\
& \multicolumn{1}{C}{GLTv1}     & 58.28 & 50.43 & 54.07 & 86.67 & \textbf{81.88} & \textbf{84.21} \\
\cdashline{2-8}[0.8pt/3pt]
& \multicolumn{1}{C}{* Baseline + \textsc{Cognisance}}  & \textbf{58.85} & 50.98 & \textbf{54.64} & \textbf{86.86} & 81.24 & 83.96 \\
& \multicolumn{1}{C}{* BLSoftmax + \textsc{Cognisance}} & 56.24 & 47.58 & 51.55 & 86.55 & 78.48 & 82.32 \\
& \multicolumn{1}{C}{* Logit-Adj + \textsc{Cognisance}} & 43.61 & \textbf{55.15} & 48.71 & 85.52 & 73.36 & 78.98 \\
\specialrule{0.6pt}{0pt}{0pt}
\multirow{4}{*}{\rotatebox[origin=c]{90}{\textbf{Augment}}}
& \multicolumn{1}{C}{Mixup}  & 55.25 & 46.76 & 50.66 & 86.95 & 82.00 & 84.40 \\
& \multicolumn{1}{C}{RandAug}       & 59.88 & 51.28 & 55.25 & 87.64 & 81.02 & 84.20 \\
\cdashline{2-8}[0.8pt/3pt]
& \multicolumn{1}{C}{* Mixup + \textsc{Cognisance}} & 64.09 & 56.57 & 60.09 & 88.81 & 81.55 & 85.03 \\
& \multicolumn{1}{C}{* RandAug + \textsc{Cognisance}} & \textbf{65.15} & \textbf{56.85} & \textbf{60.72} & \textbf{89.12} & \textbf{84.02} & \textbf{86.49} \\
\specialrule{0.6pt}{0pt}{0pt}
\multirow{4}{*}{\rotatebox[origin=c]{90}{\textbf{Ensemble}}}
& \multicolumn{1}{C}{TADE}  & 55.21 & 46.52 & 50.49 & 86.57 & 79.91 & 83.10 \\
& \multicolumn{1}{C}{RIDE}       & 60.17 & 48.28 & 53.58 & 88.03 & 81.79 & 84.80 \\
\cdashline{2-8}[0.8pt/3pt]
& \multicolumn{1}{C}{* TADE + \textsc{Cognisance}} & 58.06 & 47.44 & 52.21 & 88.00 & 81.72 & 84.74 \\
& \multicolumn{1}{C}{* RIDE + \textsc{Cognisance}} & \textbf{62.11} & \textbf{50.80} & \textbf{55.89} & \textbf{89.02} & \textbf{81.85} & \textbf{85.28} \\
\hline
\end{tabular}}
\end{table*}

\begin{figure}[htbp]
  \centering

  \subfloat[\scriptsize{CLT protocol, ImageNet-GLT}]{\includegraphics[width=0.24\textwidth]{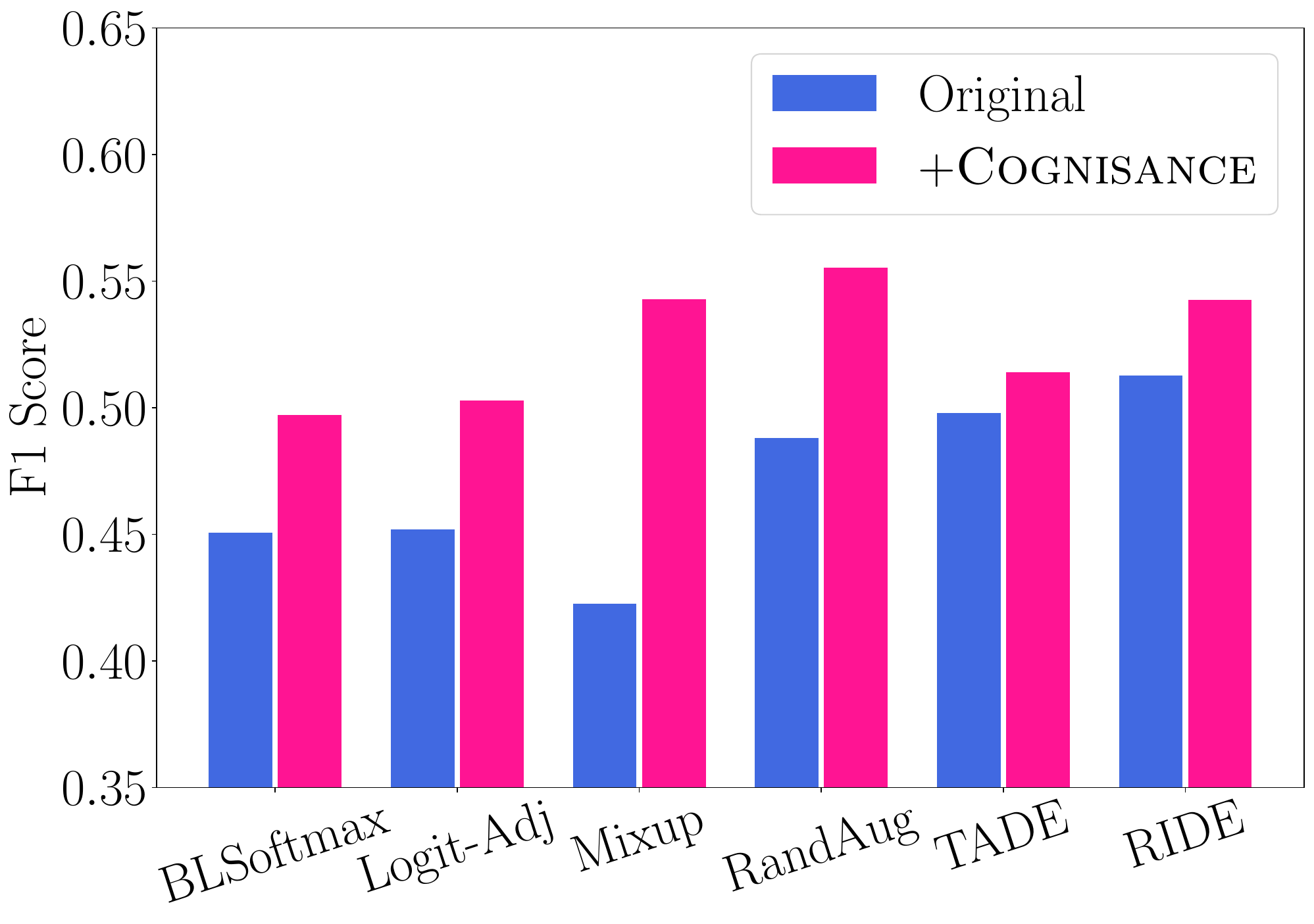}}
  \subfloat[\scriptsize{GLT protocol, ImageNet-GLT}]{\includegraphics[width=0.24\textwidth]{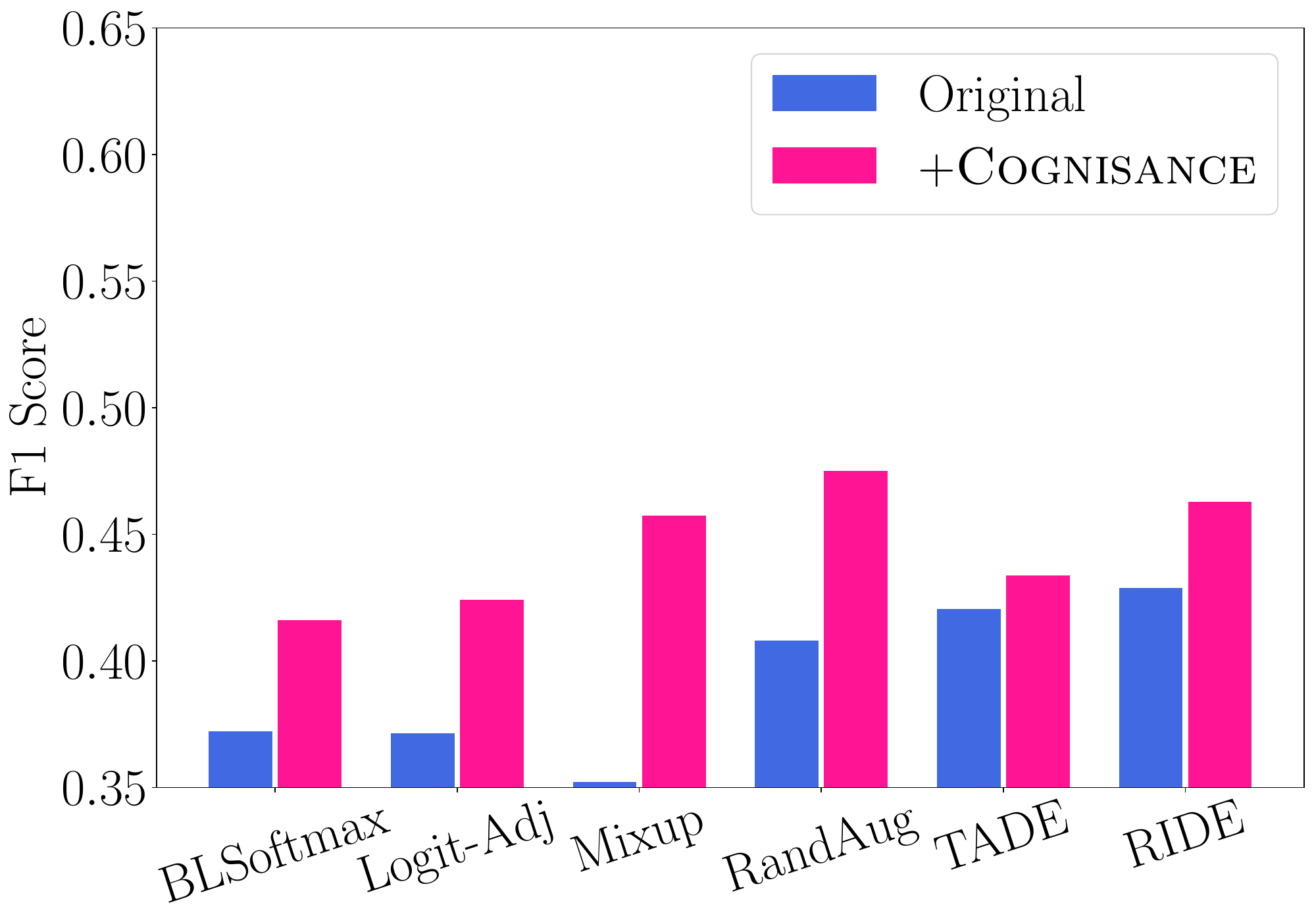}}

  \vspace{0.2cm}

  \subfloat[\scriptsize{CLT protocol, MSCOCO-GLT}]{\includegraphics[width=0.24\textwidth]{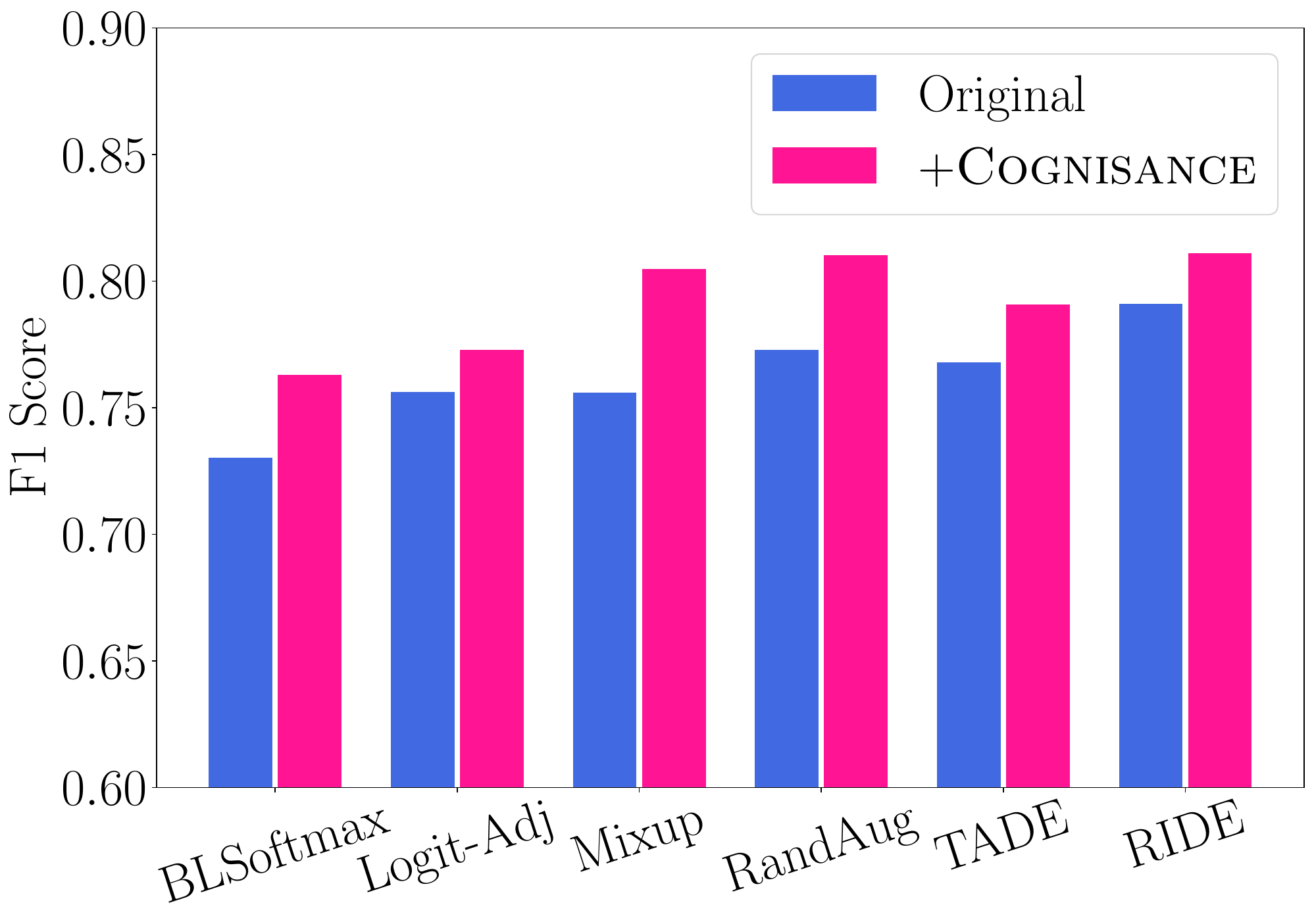}}
  \subfloat[\scriptsize{GLT protocol, MSCOCO-GLT}]{\includegraphics[width=0.24\textwidth]{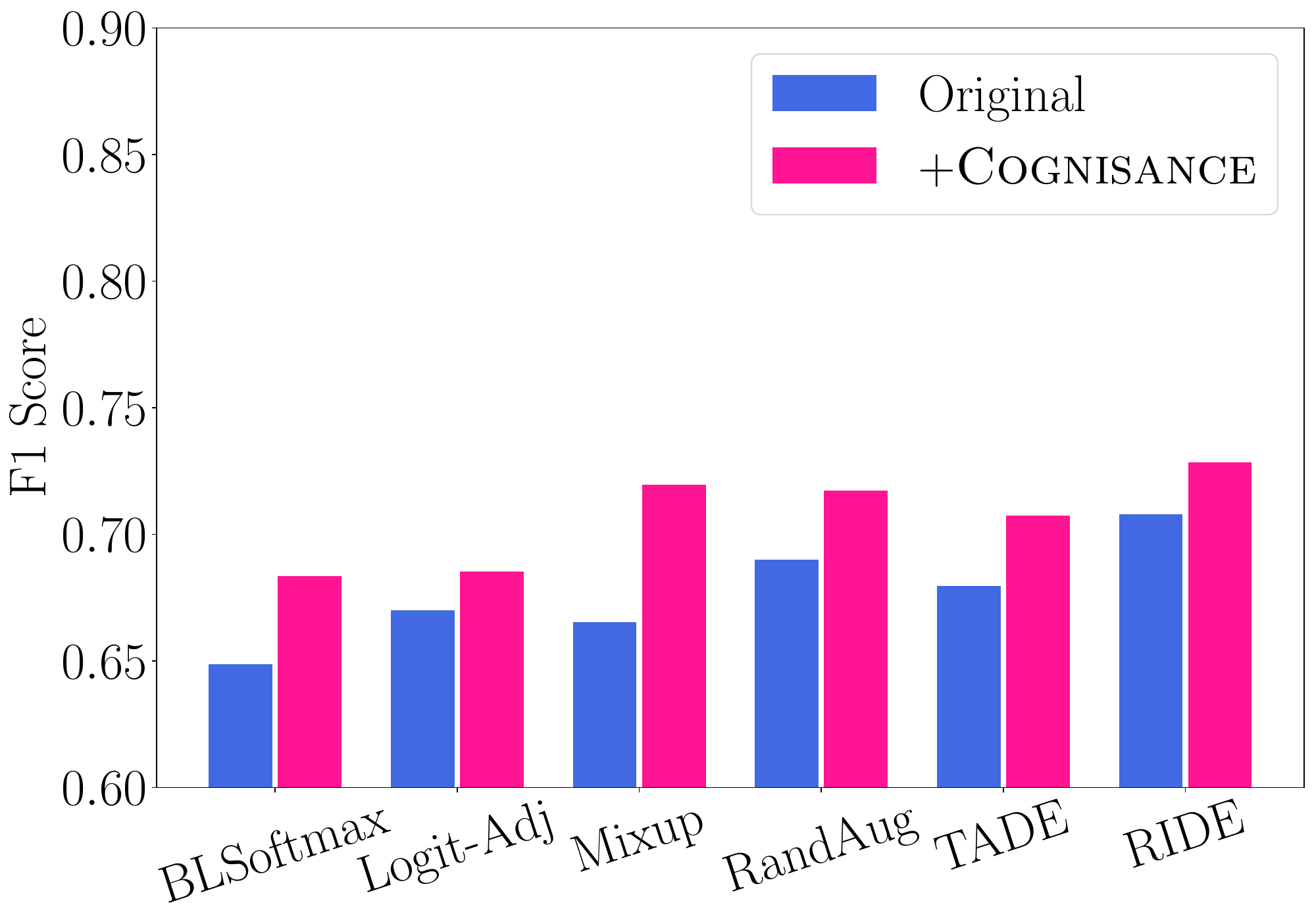}}

  \caption{\textsc{Cognisance} enhancements to existing LT methods}
  \label{fig:improvement}
\end{figure}

\textsc{Cognisance} is able to deal with both attribute-wise long-tailed and class-wise long-tailed, although the starting point of this method is to solve the problem of attribute long-tailed within the class. It meets the challenge of class-wise imbalance by eliminating the false correlation caused by long-tailed attributes \cite{tang2022invariant}.
Fig. \ref{fig:improvement} shows the improvements \textsc{Cognisance} achieved over the existing LT approaches w.r.t. the two sampling protocols (CLT and GLT) for the two benchmarks (ImageNet-GLT and COCO-GLT). One can see that the performance of all the methods has been degraded when sampling protocol is transferred from CLT to GLT, which demonstrates that the long-tailed problem is not purely class-wise but also attribute-wise (which is even more challenging). \textsc{Cognisance} can improve the performance of all existing popular LT methods on all protocols and benchmarks with an average increase up to 5\%.




Finally, Table \ref{tab:result_lt} records the experimental results of various methods on the test set of long-tailed distribution, which is consistent with the distribution of the training set. It can be seen that compared with other methods, our method achieves very encouraging results on all evaluation metrics.

{\color{black}
\subsection{Parameter Sensitivity Analysis}
Two parameters are used in the construction of CLF: $d_{rn}$ and $d_{rd}$. The parameter $d_{rn}$ represents the radius of the coarse-grained node, while $d_{rd}$ is the radius for density calculation and it can influence the fineness of tree splitting. To facilitate parameter tuning in the experimental session, we introduced the concept of the base distance. The base distance is calculated as the average distance of all sample points to their three nearest neighbors. Then, we use the base distance as the reference value for heuristic parameter tuning, which means setting $d_ {rn}$ and $d_ {rd}$ as multiples of the base distance instead of blindly adjusting them. As shown in Fig. \ref{fig:param_sensit}, we fix one parameter and adjust the other parameter. The experimental results indicate that when the parameters are set within a reasonable range, the model performance only exhibits small fluctuations. This indicates that the model is robust to changes in these parameters and can maintain stable performance over a wide range.

\begin{figure}[htbp]
  \centering

  \subfloat[\scriptsize{Adjustment to $d_ {rn}$}]{\includegraphics[width=0.24\textwidth]{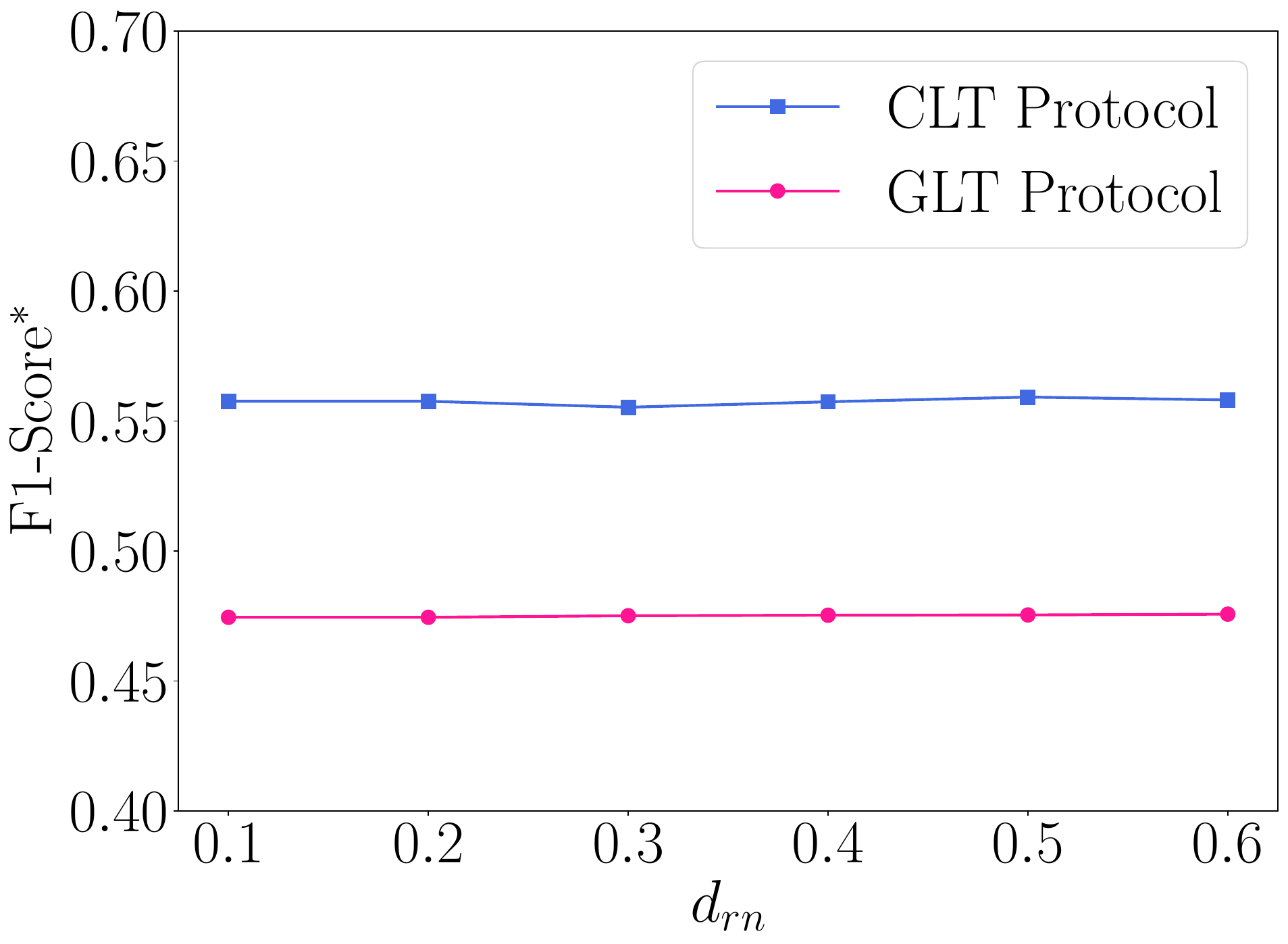}}
  \hspace{0.01cm}
  \subfloat[\scriptsize{Adjustment to $d_ {rd}$}]{\includegraphics[width=0.24\textwidth]{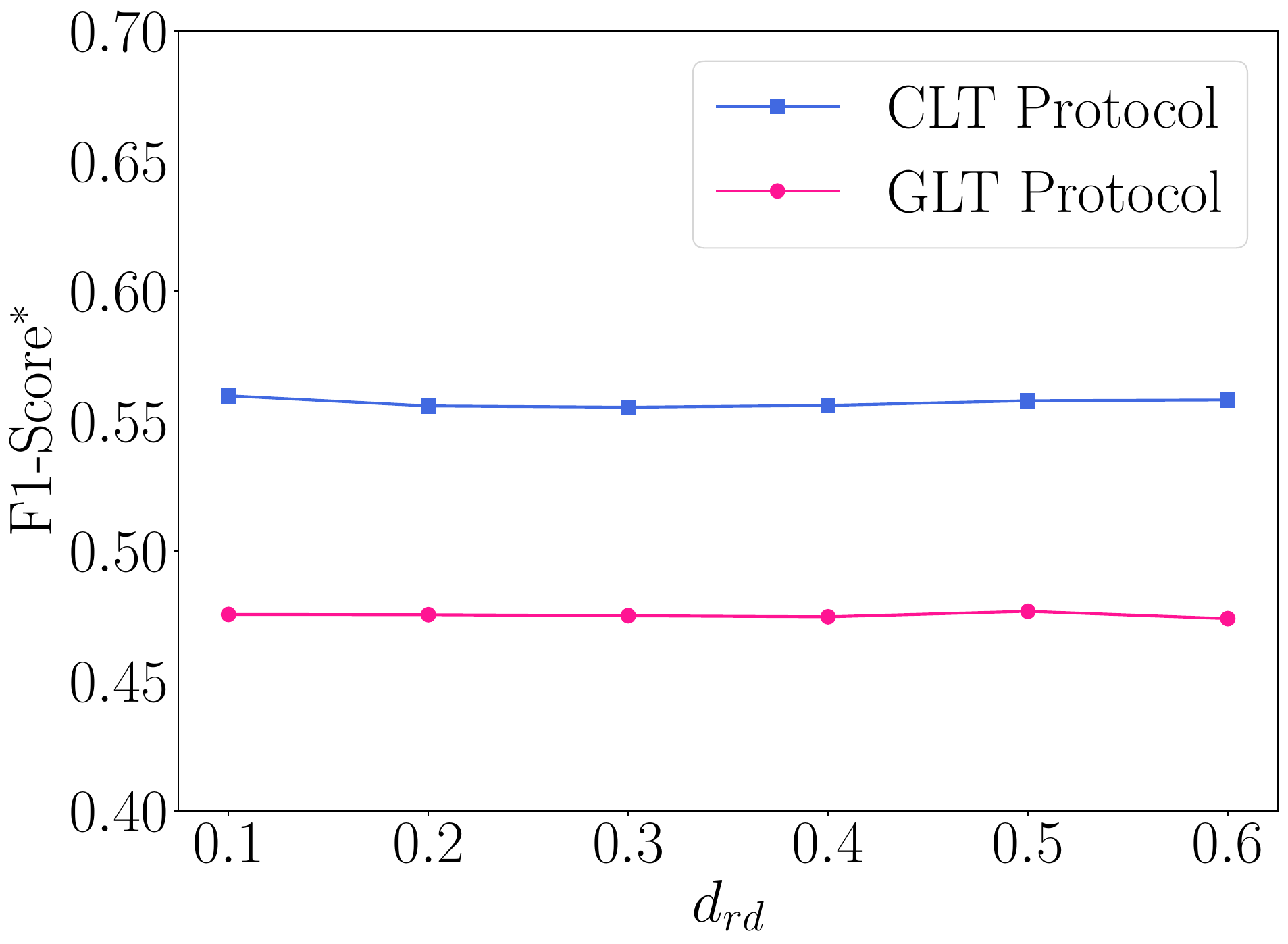}}

  \caption{{\color{black}Sensitivity analysis of parameters $d_{rn}$ and $d_{rd}$ on ImageNet-GLT, the units of horizontal and vertical coordinates are multiples of the base distance.}}
  \label{fig:param_sensit}
\end{figure}
}

{\color{black}
\subsection{Comparisons with LT Line-up on Noisy Datasets}

To test the effectiveness of the noise identification algorithm proposed in this paper, we constructed two long-tailed datasets with noise. These datasets are sampled from Animal-10N and Food-101N, both following a power-law distribution, and the imbalance ratios are 8 and 5, respectively. Detailed parameters of these datasets are shown in Table \ref{tab:info_dataset}.

\begin{table}[htbp]
    \centering
    \caption{\label{tab:info_dataset}Detailed information of Animal-10NLT and Food-101NLT datasets}
    \renewcommand{\arraystretch}{1.5}
    \resizebox{0.45\textwidth}{!}{
    \begin{tabular}{ccccccccc}
    \specialrule{1.5pt}{0pt}{0pt}
    Dataset Name & \#Training & \#Test & \#Classes & Noise Rate (\%)\\
    \hline
    Animal-10NLT & 19,261 & 5,000 & 10 & $\simeq$8 \\
    Food-101NLT & 19,230 & 3,824 & 101 & $\simeq$20 \\
    Animal-10N & 50,000 & 5,000 & 10 & $\simeq$8 \\
    Food-101N & 52,868 & 3,824 & 101 & $\simeq$20 \\
    \specialrule{1.5pt}{0pt}{0pt}
    \end{tabular}}
\end{table}

\begin{table*}[htbp]
    \newcolumntype{C}{c!{\vline width 0.75pt}}
    \caption{\label{tab:result_nlt}Performance comparison of various methods on Animal-10NLT and Food-101NLT datasets}
    \centering
    \renewcommand{\arraystretch}{1.2}
    \resizebox{0.85\textwidth}{!}{
    \begin{tabular}{c|CccCccc}
    \specialrule{1.5pt}{0pt}{0pt}
    \multicolumn{2}{C}{Dataset} & \multicolumn{3}{C}{\textbf{Animal-10NLT}} & \multicolumn{3}{c}{\textbf{Food-101NLT}} \\ \hline
    \multicolumn{2}{C}{Metric}
    & Accuracy & Precision & $\text{F1-Score}^*$ & Accuracy & Precision & $\text{F1-Score}^*$ \\
    \hline

    \multirow{7}{*}{\rotatebox[origin=c]{90}{Long-Tail Classification}}
    & \multicolumn{1}{C}{Baseline}                           & 58.70 & 61.64 & 60.14      & 21.30  & 22.49 & 21.88 \\
    & \multicolumn{1}{C}{BLSoftmax}    & 59.72 & 60.59 & 60.15      & 20.02 & 22.17 & 21.04 \\
    & \multicolumn{1}{C}{Logit-Adj}    & 60.04 & 60.76 & 60.40      & 21.94 & 21.56 & 21.75 \\
    & \multicolumn{1}{C}{Mixup}        & 61.94 & 67.98 & 64.82      & 27.02 & 30.86 & 28.81 \\
    & \multicolumn{1}{C}{RandAug} & 67.18 & 69.40 & 68.27      & 29.09 & 30.71 & 29.87 \\
    & \multicolumn{1}{C}{TADE}           & 63.80 & 64.27 & 64.03      & 26.28 & 30.28 & 28.14 \\
    & \multicolumn{1}{C}{RIDE}            & 70.52 & 70.01 & 70.26      & 28.62 & 30.12 & 29.35 \\

    \cdashline{1-8}[0.8pt/3pt]
    \multirow{4}{*}{\rotatebox[origin=c]{90}{Noise Learning}}
    & \multicolumn{1}{C}{Co-Teaching}       & 55.65 & 59.42 & 57.48       & 17.77 & 16.10 & 16.90 \\
    & \multicolumn{1}{C}{Co-Teaching$^+$}  & 56.19 & 54.23 & 55.19       & 9.96 & 6.01 & 7.5 \\
    & \multicolumn{1}{C}{RandAug + Co-Teaching}             & 57.03 & 62.33 & 59.56       & 17.8 & 18.12 & 17.96 \\
    & \multicolumn{1}{C}{RandAug + Co-Teaching$^+$}         & 51.93 & 41.68 & 46.24       & 8.54 & 3.82 & 5.28 \\

    \cdashline{1-8}[0.8pt/3pt]
    \multirow{6}{*}{\rotatebox[origin=c]{90}{Our Algorithm}}
    & \multicolumn{1}{C}{* RandAug + \textsc{Cognisance}}      & 68.32  & 71.70 & 69.97       & 30.56 & 35.70 & 32.93 \\
    & \multicolumn{1}{C}{* Baseline + \textsc{Cognisance}$^+$}  & 69.05  & 72.50 & 70.72       & 31.78 & 37.12 & 34.22 \\
    & \multicolumn{1}{C}{* BLSoftmax + \textsc{Cognisance}$^+$} & 70.12  & 72.83 & 71.45       & 30.94 & 35.84 & 33.25 \\
    & \multicolumn{1}{C}{* Logit-Adj + \textsc{Cognisance}$^+$} & 71.30  & 73.55 & 72.41       & 31.25 & 36.11 & 33.50 \\
    & \multicolumn{1}{C}{* TADE + \textsc{Cognisance}$^+$}      & \textbf{73.82} & \textbf{74.70} & \textbf{74.26}  & 33.02 & 37.78 & 35.23 \\
    & \multicolumn{1}{C}{* RIDE + \textsc{Cognisance}$^+$}      & 73.20  & 74.12 & 73.65       & \textbf{33.46} & \textbf{38.35} & \textbf{35.73} \\
    \specialrule{1.5pt}{0pt}{0pt}
    \end{tabular}}
\end{table*}

\textsc{Cognisance}$^+$ is based on the \textsc{Cognisance} framework and is decoupled from specific model structures, allowing seamless integration with other long-tail classification methods. In the comparative experiments below, we follow the experimental settings of the previous section, and a total of six from the current long-tailed methods are selected as comparisons. In the Rebalancing category, we chose BLSoftmax \cite{ren2020balanced} and Logit-Adj \cite{cao2019learning}. For the Module Improvement category, we selected RIDE \cite{wang2020long} and TADE \cite{zhang2022self}, both current SOTA methods using ensemble learning. In the Data Augmentation category, we chose Mixup \cite{zhang2017mixup} and RandAug \cite{cubuk2020randaugment}. Additionally, we included two representative methods from the noise learning domain, Co-Teaching \cite{han2018co} and Co-Teaching$^+$ \cite{yu2019does}, for comparison. As shown in Table \ref{tab:result_nlt}, methods with an asterisk (*$\langle $Method$\rangle$ + \textsc{Cognisance}$^+$) indicate those integrated with the framework proposed in this chapter, with bold numbers representing the best results across all methods. It is evident that the framework proposed in this chapter achieves the best results across all metrics.

Compared to the \textsc{Cognisance} framework proposed in the previous chapter, the \textsc{Cognisance}$^+$ framework shows significant performance improvement on the Food-101NLT dataset, which has high noise rates and fewer samples per class. Additionally, we observed that the Co-Teaching series performs worse than the Baseline on long-tail noisy datasets. This is because its noise filtering method relies on confidence, and samples from tail classes, due to their rarity, usually have low confidence, leading to a large number of misclassifications.

Since our framework already incorporates the RandAug method, we choose not to integrate it with other data augmentation methods like Mixup. Moreover, Co-Teaching methods require maintaining two models simultaneously, and our experimental setup also necessitates maintaining two models. Although our framework is capable of integrating with the Co-Teaching series, we decided against it to avoid overly complex models. Furthermore, we found convergence issues with the Co-Teaching$^+$ method on the Food-101NLT datasets. Despite multiple random seed adjustments, its performance remained poor. Since Co-Teaching$^+$ is based on the Co-Teaching method, and considering that Co-Teaching's performance on these datasets, though low, is within a normal range, we speculate that this issue may be related to the ``disagreement'' samples training used by Co-Teaching$^+$. When the performance of the two original Co-Teaching models is low, the two teacher models may have reached a consensus in their errors.}


\section{Dicussion and Further Analyses}\label{sec:Discussion}

\subsection{Why not directly apply the OLeaF?}
In \textsc{Cognisance}, CLF is more appropriate compared to OLeaF for two very important characteristics: 1) \underline{Automatic tree-splitting mechanism.} The number of parameters of CLF is relatively small, and the fine degree of tree-splitting can be controlled only by adjusting the parameter $d_{rd}$. The tree-splitting scheme in OLeaF is more rigorous and detailed but requires manually tuning parameters $d_c$ and number of trees. However, the feature learner needs to iterate in this framework and it may be possible to re-trigger clustering after each iteration. Besides, the target datasets for \textsc{Cognisance} are often large image data, so a fully automated scheme is compulsory; 2) \underline{Computing efficiency and representativeness.} Each node of CLF is a coarse-grained node because the head attribute may contain a large number of similar samples. If left un-merged, the length of the path occupied by the head attribute may skyrocket. While each node in the same path is sampled with the same weight, a large number of extremely similar nodes actually lose their representativeness. At the same time, the sampling weights of bottom nodes are compromised. So it is necessary to carry out a small range of clustering to be controlled, and this clustering process is mainly controlled by the parameter $d_{rn}$. All the member nodes within a coarse node will equally share the node's sampling weight.

\subsection{Why not directly apply IRM loss?}
In this scheme, there are two reasons why IRM loss is not used directly: 1) the original IRM loss has convergence issue on real-world large-scale image datasets; 2) the core in Multi-Center Loss lies in the Multi-center, which is a mechanism that can make the model's learning more robust because the samples within the same category in an artificial dataset may diverge tremendously. At this time, if we simply add the regularization for one center, it will impair the learning of the features instead.



\subsection{About distance measure}
The distance metric is used in two places: 1) \underline{CLF construction.} The distance matrix needs to be calculated in CLF construction, and the distance metric here can be switched arbitrarily; 2) \underline{Multi-Center Loss}. The distance from the samples to their corresponding centers will be calculated when optimizing MCL, and the distance metric here is usually consistent with that of CLF construction. It can also be switched freely. Euclidean distance is used as the default distance metric in this paper, but switching other metrics may produce better results. Due to the relatively heavy burden of computation, only one metric is evaluated here.


\section{Conclusion}\label{sec:Conclusion}
In this study, we provide insights into the long-tailed problem at two levels of granularity: class-wise and attribute-wise, and propose two important components: the CLF (Coarse-Grained Leading Forest) and the MCL (Multi-Center Loss). The CLF, as an unsupervised learning methodology, aims to capture the distribution of attributes within a class in order to lead the construction of multiple environments, thus supporting the invariant feature learning. Meanwhile, MCL, as an evolutionary version of center loss, aims to replace the traditional IRM loss to further enhance the robustness of the model on real-world datasets.

Through extensive experiments on existing benchmarks MSCOCO-GLT and ImageNet-GLT, we exhaustively demonstrate the significant improvements brought by our method. Finally, we would also like to emphasize the advantages of the two components. That is, CLF and MCL are designed as low-coupling plugins, thus can be organically integrated with other long-tailed classification methods and bring new opportunities for improving their performance.

{\color{black}Finally, in order to reduce the impact of noise samples in the long-tailed dataset on model training, we proposed \textsc{Cognisance}$^+$ framework based on \textsc{Cognisance}. In \textsc{Cognisance}$^+$, we designed an iterative noise selection scheme based on CLF. The experimental results on the Animal-10NLT and Food-101NLT datasets show that \textsc{Cognisance}$^+$ can achieve better performance  than mere Cognisance and other counterparts.}


\bibliographystyle{ieeetr}
\bibliography{Cognisance}

\begin{IEEEbiography}
[{\includegraphics[width=1in,height=1.25in,clip,keepaspectratio]{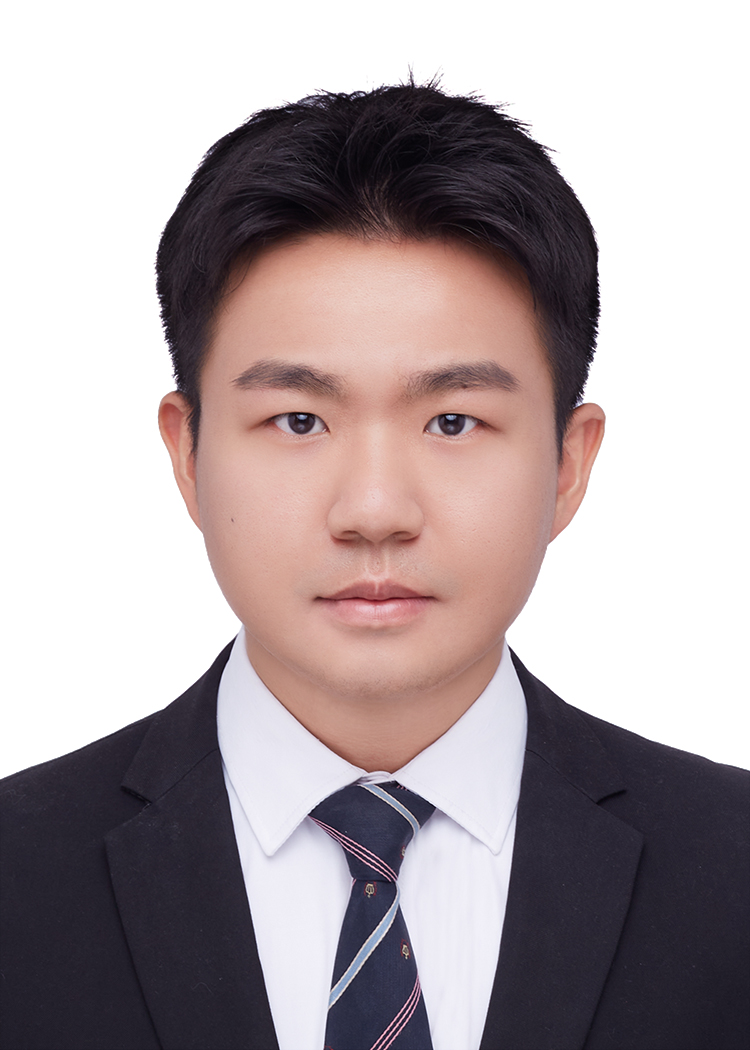}}]{Jinye Yang}
received the B.E degree from Yanshan University, Qinhuangdao, China in 2019. He is currently a graduate student with the State Key Laboratory of Public Big Data, Guizhou University. His current research interests include granular computing and machine learning.
\end{IEEEbiography}

\begin{IEEEbiography}
[{\includegraphics[width=1in,height=1.25in,clip,keepaspectratio]{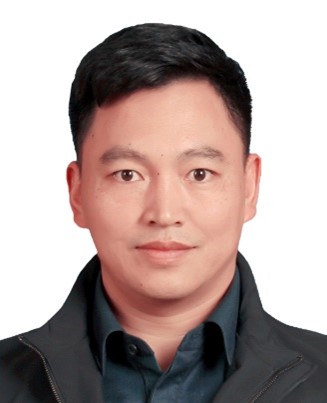}}]{Ji Xu} (M'22) received the B.S. from Beijing Jiaotong University in 2004 and the Ph.D. from Southwest Jiaotong University,
Chengdu, China, in 2017, respectively. Both degrees are in the major of Computer Science.
Now he is an associate professor with the State Key Laboratory of Public Big Data, Guizhou University.  His research interests include data mining, granular computing and machine learning. He has authored and co-authored one book and over 20 papers in refereed international journals such as IEEE TCYB, Information Sciences, Knowledge-Based Systems and Neurocomputing. He serves as a reviewer of the prestigious journals of IEEE TNNLS, IEEE TFS, IEEE TETCI, and IJAR, etc. He is a member of IEEE, CCF and CAAI.
\end{IEEEbiography}

\begin{IEEEbiography}
[{\includegraphics[width=1in,height=1.25in,clip,keepaspectratio]{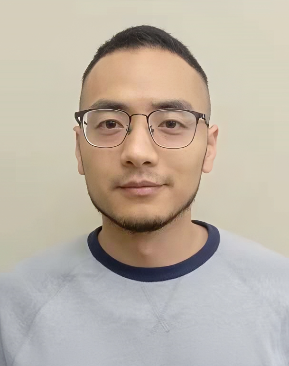}}]{Di Wu} (M') received his Ph.D. degree from the Chongqing Institute of Green and Intelligent Technology (CIGIT), Chinese Academy of Sciences (CAS), China in 2019 and then joined CIGIT, CAS, China. He is currently a Professor of the College of Computer and Information Science, Southwest University, Chongqing, China. He has over 80 publications, including 20 IEEE/ACM Transactions papers on T-KDE, T-SC, T-NNLS, T-SMC, etc., and several conference papers on ICDM, AAAI, WWW, IJCAI, ECML-PKDD, etc. His research interests include machine learning and data mining.
\end{IEEEbiography}

\begin{IEEEbiography}
[{\includegraphics[width=1in,height=1.25in,clip,keepaspectratio]{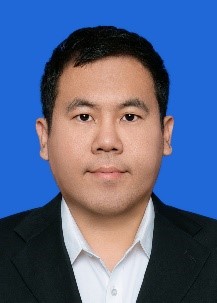}}]{Jianhang Tang} (M'23) is currently a Professor at the State Key Laboratory of Public Big Data, Guizhou University, Guiyang, China. From 2021 to 2022, he was a Lecture at the School of Information Science and Engineering, Yanshan University, Qinhuangdao, China. He has published more than 30 research papers in leading journals and flagship conferences, such as IEEE TCC, IEEE TNSM, IEEE Network, and IEEE IoTj, where two of them are ESI Highly Cited Papers. His research interests include UAV-assisted edge computing, edge intelligence, and Metaverse.
\end{IEEEbiography}

\begin{IEEEbiography}[{\includegraphics[width=1in,height=1.25in,clip]{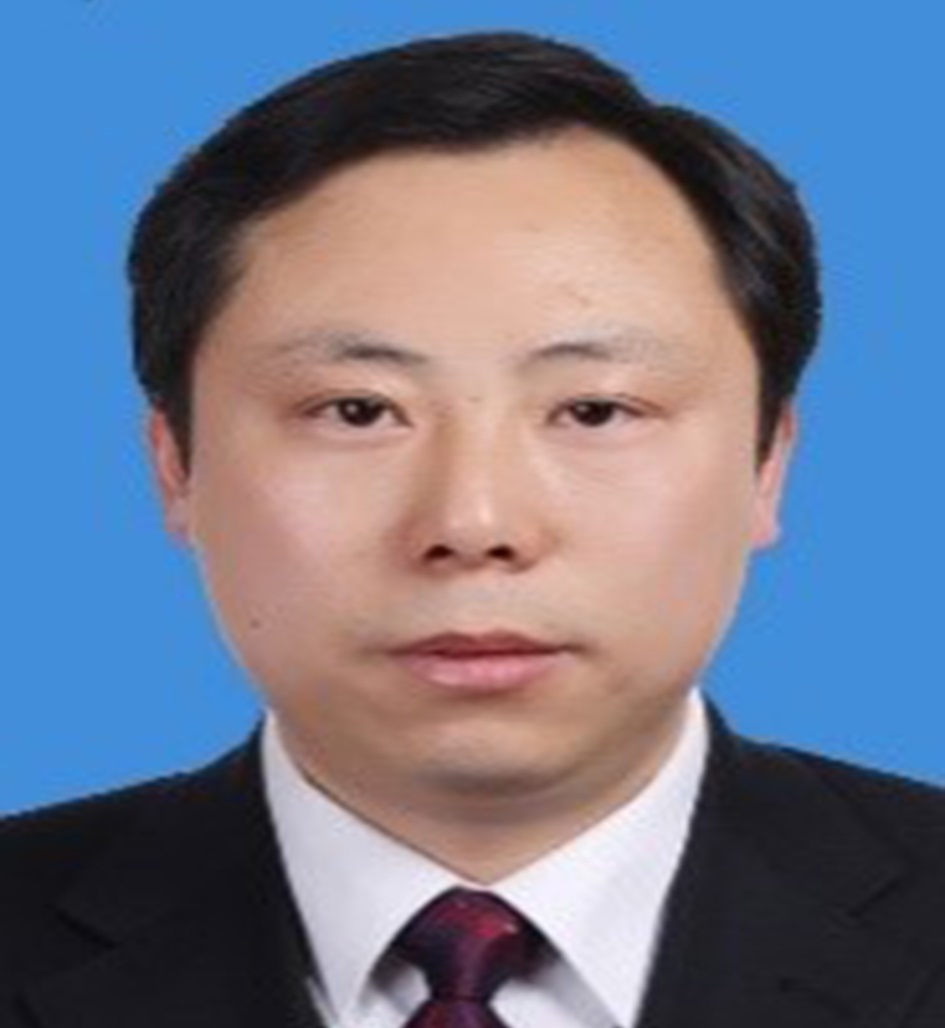}}]
{Shaobo Li}
received the Ph.D. degree in computer software and theory from the Chinese Academy of Sciences, China, in 2003. From 2007 to 2015, he was the Vice Director of the Key Laboratory of Advanced Manufacturing Technology of Ministry of Education, Guizhou University (GZU), China. He is currently the director of the State Key Laboratory of Public Big Data, GZU. He is also a part-time doctoral supervisor with the Chinese Academy of Sciences. He has published more than 200 papers in major journals and international conferences. His current research interests include big data of manufacturing and intelligent manufacturing.

\end{IEEEbiography}

\vspace{-20pt}
\begin{IEEEbiography}[{\includegraphics[width=1in,height=1.25in,clip]{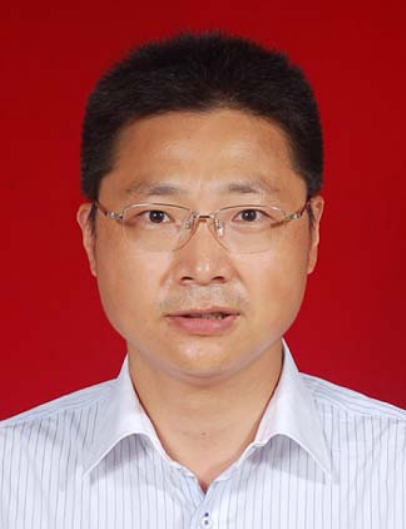}}]
{Guoyin Wang} (SM'03)
 received the B.S., M.S., and Ph.D. degrees from Xi’an Jiaotong University, Xi’an, China, in 1992, 1994, and 1996, respectively. He worked at the University of North Texas, USA, and the University of Regina, Canada, as a visiting scholar during 1998-1999. Since 1996, he has been working at the Chongqing University of Posts and Telecommunications, where he is currently the vice-president of the university. He was the President of International Rough Sets Society (IRSS) 2014-2017. He is the Chairman of the Steering Committee of IRSS and the Vice-President of Chinese Association of Artificial Intelligence. He is the author of 15 books, the editor of dozens of proceedings of international and national conferences, and has over 200 reviewed research publications. His research interests include rough set, granular computing, knowledge technology, data mining, neural network, and cognitive computing. He is a Fellow of CAAI, CCF and IRSS.

\end{IEEEbiography}

\end{document}